\documentclass[manuscript,screen]{acmart}
\AtBeginDocument{%
  }

\setcopyright{none}
\copyrightyear{This is a preprint under review}
\acmDOI{}

\usepackage{subcaption}
\usepackage[table]{xcolor}
\usepackage{multirow}
\usepackage{siunitx}
\usepackage{listings}
\usepackage{tcolorbox}
\definecolor{color1}{HTML}{FFCCCC} 
\definecolor{color2}{HTML}{CCE5FF} 
\definecolor{color3}{HTML}{D4FFCC} 
\definecolor{color4}{HTML}{FFF5CC} 
\newcommand{\FI}{FI} 
\newcommand{\SD}{SD} 
\newcommand{\CRF}{CRF} 
\newcommand{\Incomp}{I} 

\usepackage{xcolor}
\usepackage{mdframed}

\lstdefinelanguage{terraform}{
  keywords={resource, provider, variable, output, module, data, locals},
  keywordstyle=\color{blue}\bfseries,
  ndkeywords={type, default, description, required, source, version, count, for_each, depends_on},
  ndkeywordstyle=\color{blue},
  sensitive=true,
  comment=[l]{\#},
  commentstyle=\color{green!40!black},
  stringstyle=\color{red},
  morestring=[b]"
}

\begin{document}

\title{IaC Generation with LLMs: An Error Taxonomy and A Study on Configuration Knowledge Injection}


\author{Roman Nekrasov}
\email{r.s.nekrasov@student.tue.nl}
\orcid{0009-0007-7349-4389}
\author{Stefano Fossati}
\authornote{Corresponding authors}
\email{s.fossati@tue.nl}
\orcid{0009-0002-7198-883X}
\author{Indika~Kumara}
\authornotemark[1]
\email{i.p.k.weerasinghadewage@tilburguniversity.edu}
\orcid{0000-0003-4355-0494}

\affiliation{%
  \institution{Jheronimus Academy of Data Science}
  \city{'s-Hertogenbosch}
  \state{North Brabant}
  \country{Netherlands}
}
\affiliation{%
  \institution{Tilburg University}
  \city{Tilburg}
  \state{North Brabant}
  \country{Netherlands}
}
\affiliation{%
  \institution{Eindhoven University of Technology}  
  \streetaddress{Groene Loper 3}
  \city{Eindhoven}
  \state{North Brabant}
  \country{Netherlands}
  \postcode{5612 AZ}
}

\author{Damian~Andrew~Tamburri}
\email{datamburri@unisannio.it}
\orcid{0000-0003-1230-8961}

\affiliation{%
  \institution{University of Sannio}  
  \city{Benevento}
  \state{Campania}
  \country{Italy}
}
\affiliation{%
  \institution{Jheronimus Academy of Data Science}
  \city{'s-Hertogenbosch}
  \state{North Brabant}
  \country{Netherlands}
}
\affiliation{%
  \institution{Eindhoven University of Technology}  
  \city{Eindhoven}
  \state{North Brabant}
  \country{Netherlands}
  \postcode{5612 AZ}
}

\author{Willem-Jan~van~den~Heuvel}
\email{w.j.a.m.vdnHeuvel@tilburguniversity.edu}
\orcid{0000-0003-2929-413X}

\affiliation{%
  \institution{Jheronimus Academy of Data Science}
  \city{'s-Hertogenbosch}
  \state{North Brabant}
  \country{Netherlands}
}
\affiliation{%
  \institution{Tilburg University}
  \city{Tilburg}
  \state{North Brabant}
  \country{Netherlands}
}
\affiliation{%
  \institution{Eindhoven University of Technology}  
  \streetaddress{Groene Loper 3}
  \city{Eindhoven}
  \state{North Brabant}
  \country{Netherlands}
  \postcode{5612 AZ}
}

\renewcommand{\shortauthors}{Nekrasov et al.}

\begin{abstract}
  Large Language Models (LLMs) currently exhibit low success rates in generating correct and intent-aligned Infrastructure as Code (IaC). This research investigated methods to improve LLM-based IaC generation, specifically for Terraform, by systematically injecting structured configuration knowledge. To facilitate this, an existing IaC-Eval benchmark was significantly enhanced with cloud emulation and automated error analysis. Additionally, a novel error taxonomy for LLM-assisted IaC code generation was developed. A series of knowledge injection techniques was implemented and evaluated, progressing from Naive Retrieval-Augmented Generation (RAG) to more sophisticated Graph RAG approaches. These included semantic enrichment of graph components and modeling inter-resource dependencies. Experimental results demonstrated that while baseline LLM performance was poor (27.1\% overall success), injecting structured configuration knowledge increased technical validation success to 75.3\% and overall success to 62.6\%. Despite these gains in technical correctness, intent alignment plateaued, revealing a "Correctness-Congruence Gap" where LLMs can become proficient "coders" but remain limited "architects" in fulfilling nuanced user intent.
\end{abstract}



\maketitle

\section{Introduction}
The rapid adoption of cloud computing has fundamentally transformed how organizations design, deploy, and manage their technological infrastructure~\cite{leite2019survey, guerriero2019adoption, kumara2021s}. This transformation has given rise to Infrastructure-as-Code (IaC), a paradigm that treats infrastructure configuration as software programs, enabling version control, automated testing, and reproducible deployments~\cite{morris2020infrastructure}. However, writing correct IaC requires expertise due to the number of configurable resources, their dependencies, schemas, and constraints~\cite{aws2024overview,nasiri2024towards}. This complexity creates a substantial barrier to entry and increases the likelihood of misconfigurations that can lead to security vulnerabilities, operational failures, or excessive costs~\cite{guerriero2019adoption, rahman2020gang}.

The emergence of Large Language Models (LLMs) has opened new possibilities for automated code generation across various programming domains~\cite{vaswani2017attention, brown2020language, chen2021evaluating}. 
Models like GPT-4~\footnote{\url{https://openai.com/index/hello-gpt-4o/}}, Codex~\cite{chen2021evaluating}, and specialized code-generating systems have demonstrated impressive capabilities in translating natural language descriptions into functional code~\cite{chen2021evaluating, brown2020language}. This success has naturally led to interest in applying LLMs to IaC generation~\cite{kon2024iac, pujar2023automated, pimparkhede2024doccgen, lee2024llm}, potentially democratizing cloud infrastructure management by allowing users to describe their needs in natural language rather than wrestling with complex configuration syntax.

Yet, initial attempts at LLM-based IaC generation have revealed fundamental challenges. Unlike traditional programming, where code can be tested through execution, IaC correctness depends on adherence to provider schemas, proper handling of resource dependencies, and alignment with operational requirements that are rarely explicit~\cite{kon2024iac, pujar2023automated, lee2024llm}. 

Recent benchmarks show that even state-of-the-art models achieve only 19-27\% success rates on IaC generation tasks, compared to over 80\% on general programming benchmarks~\cite{kon2024iac}. This performance gap highlights the distinct challenges of the IaC domain and the necessity for specialized approaches.

To study and evaluate an LLM-based approach for IaC generation, we formulate the following research questions:

\begin{itemize}
    \item \textbf{RQ1}: \textbf{What are the characteristics and possible causes of errors made by LLMs when generating IaC in Terraform?}
    Understanding failure patterns is essential for designing targeted solutions. Without a systematic analysis of how and why LLMs fail at IaC generation, improvement strategies remain unfocused. We consider Terraform as it is one of the main IaC frameworks used by both academia and industry. 
    \item \textbf{RQ2:} \textbf{How can configuration knowledge improve the overall quality of LLM-generated Terraform code?} 
    Given the identified error patterns, this question investigates whether and how structured configuration knowledge injection can address the root causes of generation failures. Particularly by comparing different forms of configuration knowledge (e.g., semantic chunks vs. a relational knowledge graph) and various graph enrichment strategies.
\end{itemize}

To answer our research questions, we first investigated the characteristics that a reliable benchmark for evaluating LLM-generated IaC must possess, starting from the IaC-Eval benchmark as a baseline~\cite{kon2024iac}. Based on our tool analysis, we propose enhancements to address each of the identified limitations. For instance, we automate the evaluation and error analysis of IaC generation pipelines, integrate cloud emulation, and enhance the pipeline's validation phase by reviewing intent specifications and correcting errors in Terraform scripts. The enhancements are described in detail, covering motivation, implementation, and benefits. 

Secondly, we performed a comprehensive analysis of error characteristics and causes across 458 IaC Terraform generation tasks. We introduced a two-dimensional taxonomy that systematically
categorizes the types of errors found in LLM-generated Terraform code. The taxonomy classifies errors by validation stage and underlying generation failure pattern. In particular, we focused on technical validation failures and intent validation, as they represent distinct challenges in IaC generation.

We investigated and compared the impact of the knowledge configuration by developing and evaluating systematic methods for injecting configuration knowledge into the generation process. Through the design and implementation of multiple knowledge representation and retrieval strategies, we investigate how structured domain knowledge can improve  Terraform code generation. In particular, we established performance baselines by comparing LLM-only without Retrieval-Augmented Generation (RAG), Naive RAG, and Graph RAG approaches, to understand how knowledge structure fundamentally affects generation quality. RAG enables LLMs to retrieve and use relevant external information to improve their generation tasks~\cite{fan2024survey}. A Graph RAG represents external information as a knowledge graph, a network of nodes (entities) and edges (relationships), enabling LLMs to perform multi-hop reasoning in the retrieval task ~\cite{edge2024local}.  In this study, we explored three targeted Graph RAG enhancement strategies — retrieval scope expansion, semantic enrichment, and dependency modeling — each designed to address specific limitations identified in the baseline analysis. 

Lastly, through systematic experimentation, we evaluated the different approaches to organizing and delivering IaC configuration knowledge, examining their impact on both technical correctness and intent alignment.  This systematic progression from baseline through targeted enhancements to adaptive mechanisms provides comprehensive insights into how configuration knowledge, knowledge structure, and injection strategies affect the quality of LLM-based IaC generation.

This article presents the following contributions:
\begin{itemize}
    \item We present the enhanced IaC-Eval framework developed for this research. We reported improvements, validation corrections, automated error analysis, and automated orchestration that enable reproducible evaluation of IaC generation approaches.
    \item We developed a comprehensive two-dimensional taxonomy of errors in LLM-generated Terraform code. Through systematic analysis of generation failures, we identified 15 key error patterns -- including factual incorrectness, incompleteness, and failures in contextual reasoning -- that motivate the knowledge injection approaches.
    \item We present the core technical contributions, including the implementation of Naive RAG and Graph RAG systems, systematic enhancements to Graph RAG (semantic matching, LLM-enhanced descriptions, and reference relationships).
    \item We provide the experimental results comparing all approaches. We demonstrate how structured knowledge representation improves overall success rates (both technical validation and intent validation) from 27.1\% to 60.4\%, while revealing persistent challenges in intent alignment that create a performance plateau despite substantial technical improvements.
    \item To ensure reproducibility and enable further research, we provide a replication package at: \url{https://figshare.com/s/31403e6aeefeb4366bf1}. The package includes the enhanced IaC-eval framework code and its related revised dataset. It also contains the complete implementation of the knowledge injection techniques, along with their experiments and the associated results. The scripts for statistical analysis are also provided in the package.
\end{itemize}

The paper is organized as follows. Section~\ref{sec:related} provides an overview of existing research in IaC LLM generation. Section~\ref{sec:benchmark} presents the baseline Terraform benchmark framework and its enhancements developed in this study. In Section~\ref{sec:error_taxonomy}, we describe the two-dimensional IaC error taxonomy. Section~\ref{sec:methods} presents the LLM knowledge injection techniques implemented for this study, and Section~\ref{sec:eval} evaluates these techniques on the error taxonomy presented in Section~\ref{sec:benchmark}. Section~\ref{sec:discuss} discusses our results. Finally, we describe the limitations and threats to the validity of our study in Section~\ref{sec:threats} and draw our conclusions in Section~\ref{chapter:conclusions}.

\section{Related Works}\label{sec:related}
Building on the general capabilities and limitations of code-generating LLMs, this section examines their specific application to IaC generation. The declarative nature of most IaC languages, combined with their domain-specific requirements, creates unique challenges for LLM-based generation approaches.

\subsection{Preliminary Work on NL2IaC}
Early research has begun to explore the generation of IaC with LLMs directly. Pujar et al.~\cite{pujar2023automated} introduced Ansible Wisdom, a system for generating Ansible YAML configurations using domain-specific LLMs. Their work demonstrated that pretraining on curated Ansible datasets and carefully engineered prompt strategies significantly improves code structure and schema correctness. They introduced novel Ansible task- and playbook-specific evaluation metrics, though they primarily relied on BLEU (Bilingual Evaluation Understudy) scores rather than on functional deployment success.

Building on the importance of schema correctness and leveraging external knowledge for domain-specific languages, Pimparkhede et al.~\cite{pimparkhede2024doccgen} introduced DocCGen, a framework for document-based controlled code generation for structured DSLs (domain-specific languages) such as Ansible YAML and Bash commands. DocCGen's approach of retrieving libraries (e.g., Ansible modules) to find relevant documentation for a given prompt, followed by constrained decoding that enforces schema and grammar rules from these documents during code synthesis, showed notable improvements. This technique was reported to reduce hallucinations and syntax errors, with early results indicating that even smaller, constrained models could outperform larger unconstrained ones, particularly on unseen modules. The authors highlighted key struggles for LLMs with DSLs, such as a lack of schema awareness, prompt sensitivity, low data availability for specific DSLs, and difficulties with complex nested structures, which their framework aims to mitigate.

Subsequently, and in line with the strategy of leveraging external information, Lee et al.~\cite{lee2024llm} proposed a broader LLM-driven framework that incorporates requirements refinement, RAG, and ensemble-based code synthesis to generate deployable infrastructure code. Their prototype system used multiple LLMs in a pipeline to iteratively refine inputs and combine outputs for generating Ansible scripts. While they demonstrated the potential to create plausible configurations such as NGINX deployments on Kubernetes, their generated scripts still often suffered from semantic hallucinations and syntax errors. 

These works highlight that LLM-based IaC generation poses unique challenges compared to traditional code generation. The declarative nature of IaC means correctness often cannot be verified through simple test case execution, and standard code evaluation metrics like BLEU or exact match are insufficient, since syntactically different scripts can be functionally equivalent~\cite{xu2024cloudeval, kon2024iac}. Furthermore, visually plausible scripts may fail during deployment due to subtle configuration errors that are difficult to detect without execution.

\subsection{Benchmarking IaC Generation: IaC-Eval}
Given the unique challenges IaC introduces, there is a need for a benchmark designed explicitly for IaC to evaluate IaC generation. Kon et al.~\cite{kon2024iac} developed the first dedicated benchmark for evaluating LLM-generated IaC programs. Focusing on Terraform configurations, IaC-Eval provides a validation approach that assesses not only syntactic correctness but also functional alignment with infrastructure intent. More specifically, IaC-Eval provided a natural language IaC problem description and its related infrastructure intent specification.

Experimental results reveal a substantial gap in current model performance. For instance, GPT-4 achieved only a 19.36\% pass@1 accuracy on IaC-Eval. This sharp contrast highlights the domain-specific difficulty of IaC tasks. The benchmark further shows that standard LLM enhancements, such as few-shot prompting, chain-of-thought reasoning, and multi-turn interaction, offer limited improvements. RAG strategies showed the most promising gains, suggesting that access to domain-specific knowledge is particularly valuable. This resonates with the emphasis on external knowledge seen in the previously described works by Lee et al.~\cite{lee2024llm} and Pimparkhede et al~\cite{pimparkhede2024doccgen}.

\subsection{Related Work on LLMs in Infrastructure as Code}
Beyond direct code generation, several complementary studies have explored adjacent challenges in applying LLMs to IaC contexts, including quality assurance, security configuration, and automated repair.

Haque et al.~\cite{haque2022kgsecconfig} introduced KGSecConfig, a knowledge graph-based system for securing container orchestrator (CO) configurations, such as Kubernetes and Docker. Although not directly aimed at generation, their system automates security configuration by structurally extracting and semantically linking arguments, options, and concepts from official documentation. This supports tasks such as mitigating misconfiguration and enforcing policies, highlighting the potential of knowledge-based reasoning in IaC environments.

Zhang et al.~\cite{zhang2024does} investigated the quality of ChatGPT-generated Kubernetes manifests and found that over 35\% of generated manifests contain configuration smells. This empirical study highlights the risks of relying solely on LLMs for IaC and the need for post-generation validation or guided synthesis techniques.

Diaz-de-Arcaya et al.~\cite{diaz2024towards} explored automated patching of IaC scripts using local LLMs. Their self-healing framework evaluates the ability of offline models to correct common Ansible script errors through role prompting and scenario-based validation. Based on preliminary results from experiments with a few models, their work suggests that local LLMs can enhance the reliability of IaC in resource-constrained environments.

Hassan et al.~\cite{hassan2024state} focused on comparing the infrastructure state before an infrastructure modification; more specifically, they investigated state reconciliation defects in IaC.  They categorized 5,110 issues from the Ansible ecosystem and developed heuristics to guide LLM-based validation. Their defect taxonomy and prompt-based validation system offer a structured foundation for integrating LLMs into IaC testing pipelines, especially for detecting misalignment between declared and actual states.

These studies collectively highlight challenges and opportunities in applying LLMs to IaC contexts, spanning generation quality, security configuration, automated repair, and defect detection.

In our study, we first aim to enhance the IaC-Eval framework and, secondly, create a taxonomy of errors that can occur in the LLM-based IaC generation. Next, we aim to examine the different types of knowledge injection methods for LLMs. By evaluating these methods specifically within Terraform-based code generation tasks, our research uniquely contributes to understanding how various types of injected knowledge in LLMs can optimize and refine IaC generation.

\section{An Open Source Benchmark for IaC Code Generation}\label{sec:benchmark}

IaC-Eval benchmark~\cite{kon2024iac} is the baseline of our study. We selected this benchmark because it focuses on Terraform code generation and considers intent-based IaC development, where IaC scripts are developed to meet high-level user intents or goals~\cite{Tamburri2019}.  Moreover, when our research project was started, IaC-Eval was the only available benchmark for IaC.  Given an infrastructure description from a dataset of intents, the IaC eval system prompts the LLM to generate Terraform IaC code. If the generated IaC is syntactically correct, it is considered technically valid. Next, IaC Eval compares the dependency graph of the generated IaC against the user-intended infrastructure specification, using Open Policy Agent (OPA)~\footnote{https://www.openpolicyagent.org/}. If the user intent is satisfied, the IaC is considered correct during intent validation. 

\subsection{Limitations of IaC-Eval Benchmark}
When we experimented with the benchmark, we observed several constraints/limitations that can hinder experimentation using it. The first author set up the benchmark and ran the experiments, and discussed the issues encountered with the second and third authors.

First, the benchmark relies on direct AWS API interaction, which adds to management complexity, such as authentication and dependency management in experimental environments, limiting its reproducibility. In addition, the continuous API communication creates a persistent risk of accidental resource provisioning during benchmark development and testing. Second, when running the provided "ground truth" Terraform scripts, only 89\% passed the \texttt{terraform init} stage, and only 59\% successfully completed the \texttt{terraform plan} validation. These failures come from syntactic, semantic, and missing package problems. Third, the benchmark's initial dependency on a specific, undocumented OPA version created reproducibility challenges. After updating to OPA 1.1.0, we still encountered failures during intent specification validation. Fourth, the accompanying GitHub repository contained unresolved dependency conflicts, outdated package references, and inadequate documentation for the environment configuration. Last, the benchmark evaluation is based on binary pass-fail metrics without illuminating the underlying failure patterns or generation limitations. To understand the nature of the failure, a manual inspection was needed.

\subsection{Enhancements and Contributions to the Benchmark}
\begin{figure}[t]
    \centering
    \begin{subfigure}[c]{0.6\textwidth}
        \centering
        \includegraphics[width=\linewidth]{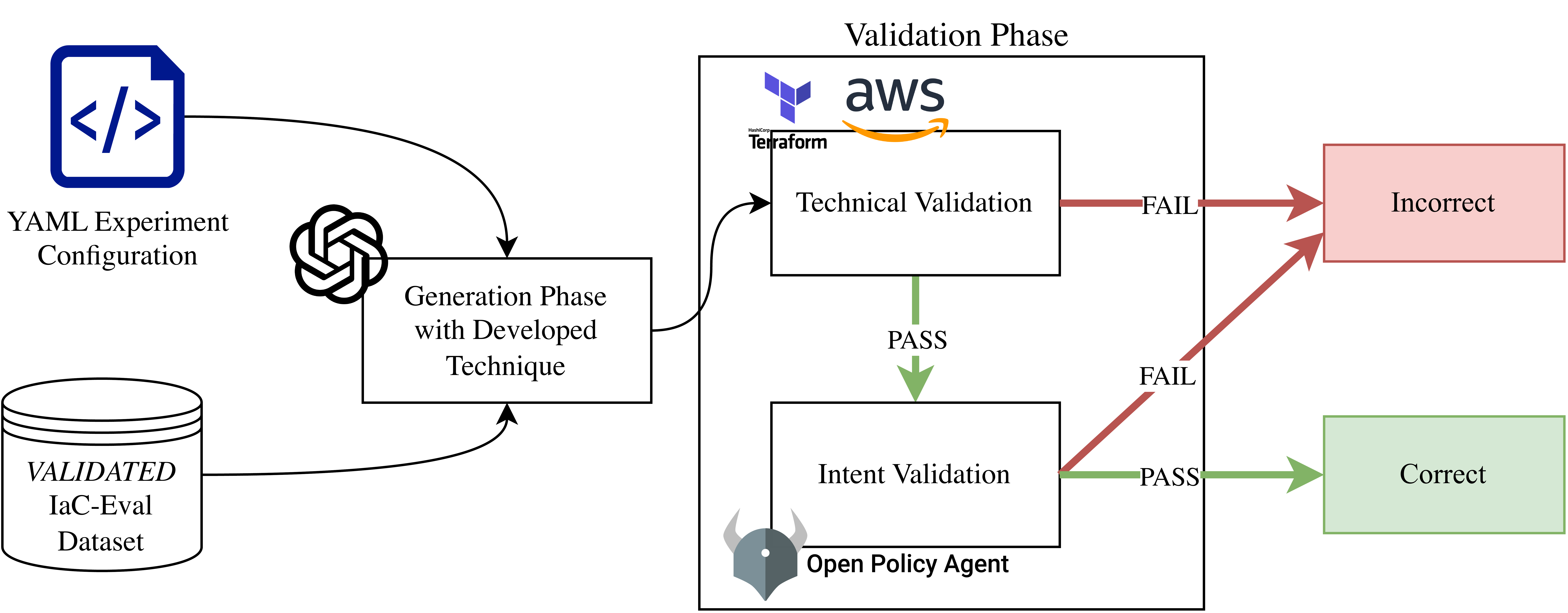}
       \caption{IaC-Eval Architecture~\cite{kon2024iac}}
       \label{fig:iac-benchmark}
    \end{subfigure}
    \hfill
    \begin{subfigure}[b]{0.8\textwidth}
        \centering
       \includegraphics[width=\linewidth]{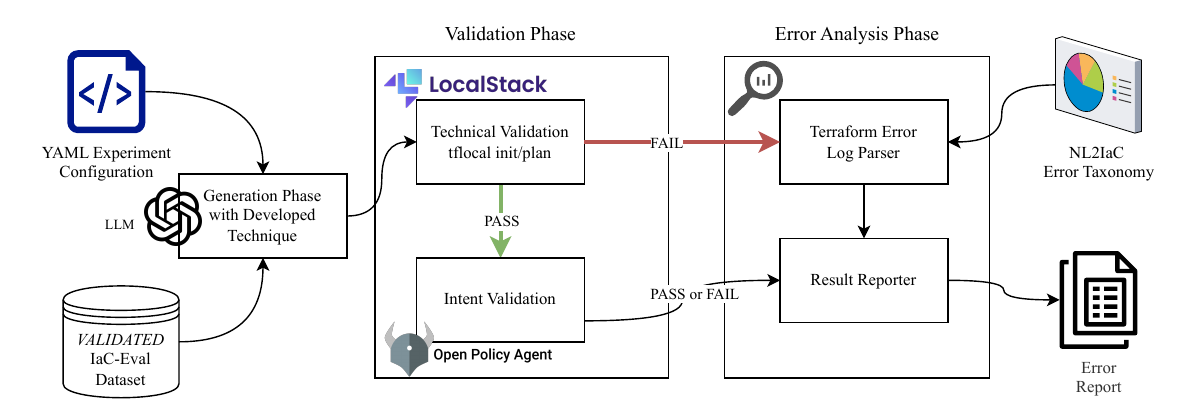}
       \caption{Enhanced IaC-Eval Benchmark Architecture}
       \label{fig:enhanced-benchmark}
    \end{subfigure}
    \caption{IaC-Eval Benchmark Comparison}
    \label{fig:iac-eval}
\end{figure}

The limitations identified in the previous section collectively motivated our redesign of the IaC-Eval framework. 
We developed an integrated solution that addresses each limitation category through architectural and implementation enhancements that collectively transform the benchmark's reproducibility, accessibility, and analytical depth. Figure~\ref{fig:iac-eval} illustrates the key differences between the original framework and the enhanced one.  Our resigned IaC-Eval consists of three main phases: IaC generation using a specific LLM-based technique, a two-stage validation pipeline, and a detailed error analysis based on the taxonomy developed in this study (see Section \ref{sec:error_taxonomy}).

\paragraph{Automated Evaluation Pipeline.}
We developed a fully automated evaluation pipeline that streamlines the entire experimental workflow, maximizing reproducibility while minimizing manual intervention. The orchestration framework requires a YAML template specifying only the generation technique and experiment name; after that, the framework automatically executes all evaluation stages with appropriate parameter propagation and result tracking. The implementation is based on Python 3, with only OPA and LocalStack as dependencies.

\paragraph{Cloud Emulation Integration.} A significant enhancement in our benchmark is the integration of LocalStack~\footnote{https://www.localstack.cloud/} as a complete replacement for real AWS services. LocalStack supports most AWS services and enables users to test and debug their AWS cloud applications locally. This change eliminates the need to create and manage AWS accounts, prevents cross-contamination between evaluation runs, and eliminates non-deterministic behavior caused by account-specific configurations or restrictions. Moreover, the emulator's local execution eliminates network latency and API rate limiting that frequently constrained the original benchmark's performance.

\paragraph{Automated Error Analysis.}
We developed an error analyzer module that processes and categorizes errors from IaC validation outputs, which enhances the original benchmark's binary pass-fail metrics, which only indicated whether the IaC code was generated successfully or not, without specifying the type of failure in case of unsuccessful code generation.  The new error classification is based on a taxonomy described in the next section. This module combines technical validation results from Terraform operations and OPA intent validation status to determine an overall status (Success/Failed) for each IaC configuration and identify the specific failure stage. The analyzer generates comprehensive statistics, including overall success/failure rates, distributions of failures by stage, counts of different error categories and subcategories (excluding OPA validation errors from general error counts), and counts of errors per script. Finally, it produces a Markdown report and detailed error logs in CSV format.

\paragraph{Validation Phase Enhancements.}
We recognized the substantial effort the original authors invested in developing the intent specifications and reference implementations, and thus focused on preserving their underlying evaluation philosophy while addressing the identified inconsistencies and technical issues. Our validation enhancement process involved manual review and correction of the intent specifications to ensure alignment with actual infrastructure requirements. We identified and rectified 117 of 458 OPA policy-as-code scripts (Rego Policy Language) that contained errors or overly strict policies that overfitted to specific implementation patterns in the reference scripts. These corrections maintained the original evaluation intent while accommodating the legitimate implementation variations that Terraform's declarative nature supports. For the reference Terraform scripts, we conducted a complete review and correction process that addressed syntax errors, missing file references, and deprecated resource configurations. This effort established a reliable baseline against which to compare generated outputs.

\section{Taxonomy of NL2IaC Errors}\label{sec:error_taxonomy}

In this section, we introduce a two-dimensional taxonomy that systematically categorizes the types of errors found in LLM-generated Terraform code, to have more insights into the nature of the failure or the LLM model behaviors that caused it than the simple pass/fail outcome from validation pipelines used in IaC-Eval~\cite{kon2024iac}. This taxonomy focuses on what the model is doing wrong, both in terms of how the error surfaces during validation and the structure of the incorrect output itself. The first dimension classifies the validation stage at which the error is detected, covering syntax violations, schema mismatches, runtime issues, and intent misalignments. These categories reflect the external, observable failures that prevent successful deployment or fulfillment of infrastructure intent.
The second dimension captures the error pattern in the generated code, describing how the output deviates from correctness. This includes patterns such as factual incorrectness (e.g., hallucinated arguments or resource types), incompleteness (e.g., missing required blocks or references), failures in contextual reasoning (e.g., incorrect dependencies or conflicting arguments), and structural deficits (e.g., syntax errors). These categories reflect the internal generation flaws that cause validation failures.

The remainder of this section describes the methodology used to construct the taxonomy, presents its detailed taxonomy, and analyzes the prevalence and distribution of error types across technical and intent validation stages. Through this, we aim to offer a diagnostic lens on LLM behavior that supports both evaluation and generation improvements for IaC.

\subsection{Error Taxonomy Development}
The development of our NL2IaC error taxonomy followed a systematic methodology that used qualitative data analysis methods (as in other taxonomy-creation studies~\cite{Rahman8812041,rahman2020gang}). This section describes the key stages of this process: initial corpus generation, systematic error data collection via our validation framework, and the multi-step qualitative coding procedure for error classification. Additionally, we collected and mined Terraform provider changelogs to contextualize temporal errors.

\subsubsection{Terraform Error Corpus}
\begin{figure}[t]
    \centering
    \begin{subfigure}[b]{0.45\textwidth}
        \includegraphics[width=\textwidth]{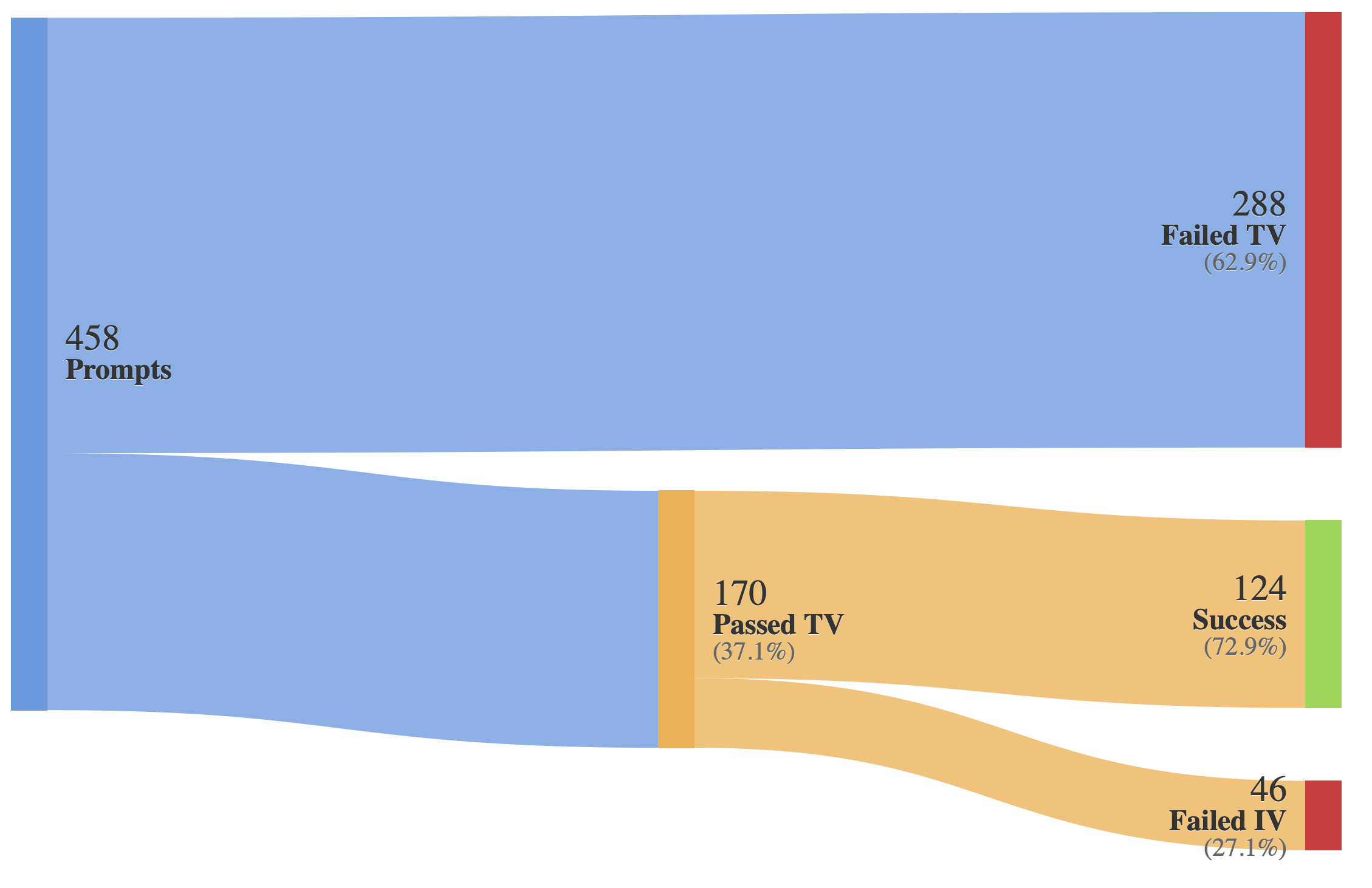}
        \caption{Flow of 458 LLM-Generated Terraform Scripts Through the Two-stage Validation Pipeline}
        \label{fig:sankey}
    \end{subfigure}
    \hfill
    \begin{subfigure}[b]{0.49\textwidth}
        \includegraphics[width=\textwidth]{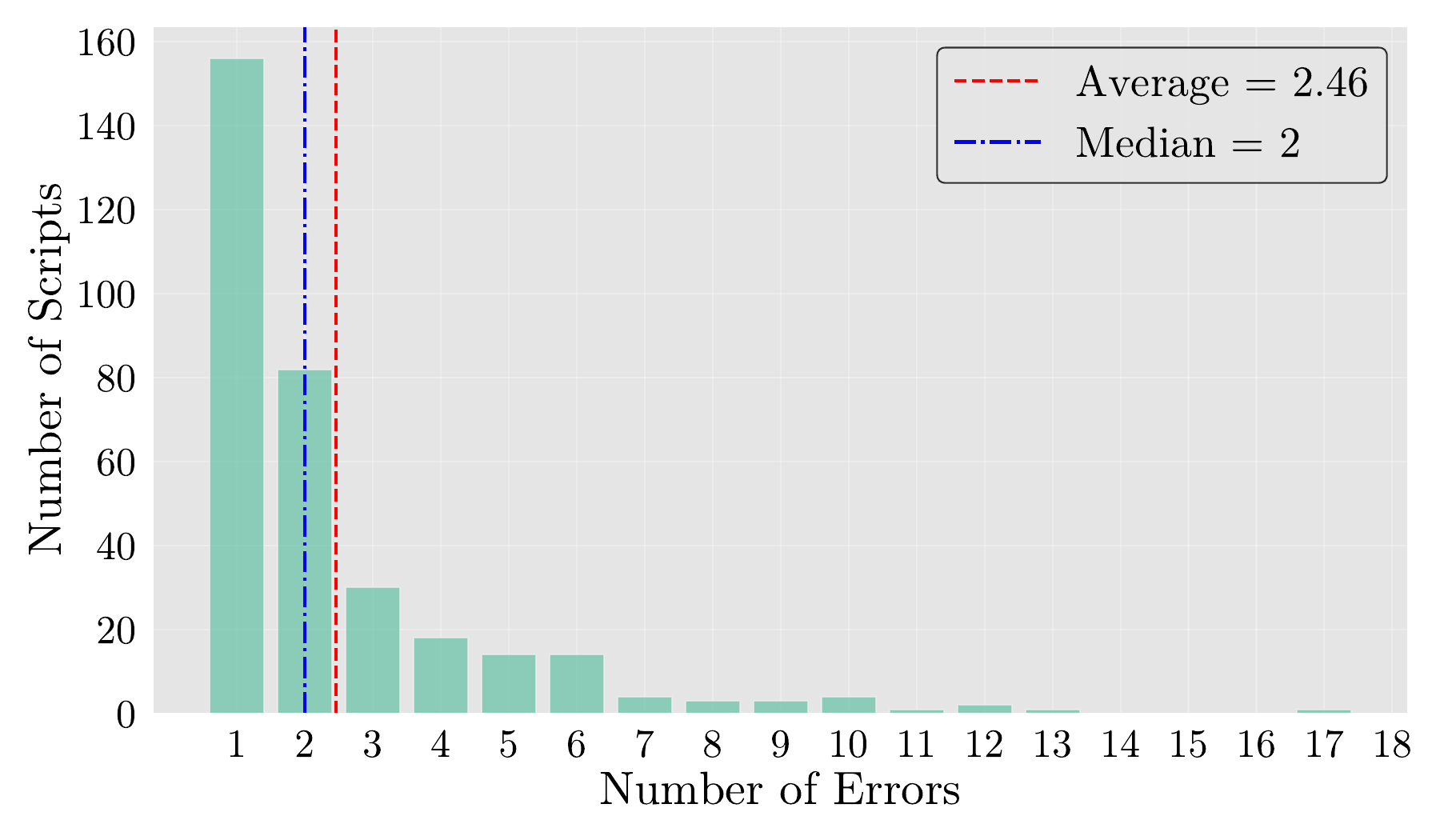}
        \caption{Distribution of Error Counts per Script (TV Failures)}
        \label{fig:error_distro}
    \end{subfigure}
    \caption{Baseline Error Statistics}
    \label{fig:baseline_stat}
\end{figure}

To construct our error corpus, we executed our enhanced validation pipeline using GPT-4o as the baseline Large Language Model (LLM), selected for its leading performance in the IaC-Eval benchmark. For each of the 458 input prompts, the generated Terraform script was processed. The output logs from Terraform's technical validation (TV) and the binary-passed/failed status from the Open Policy Agent (OPA) intent verification (IV) for each script were systematically stored. As visualized in Figure~\ref{fig:sankey}, a significant portion of the initial 458 scripts, 288 scripts (62.9\%), failed the initial technical validation phase. The remaining 170 scripts (37.1\%) successfully passed technical validation; of these, 124 (27.1\% of the total) also passed intent validation, while 46 (10.0\% of the total) failed the intent validation stage. This process yielded 774 unique error log entries from the technical validation phase and identified 46 scripts with intent validation failures.

Figure~\ref{fig:error_distro} illustrates the distribution of errors per script.  The histogram indicates that it is common for a single script to contain multiple errors. The average number of errors per failed script is 2.46, with a median of 2 errors. While many scripts have only one or two errors, some contain significantly more, with a maximum of 17 errors in a single script.

\subsubsection{Coding Procedure}
After collecting the dataset of LLM-generated Infrastructure as Code (IaC) and the associated validation logs, we followed a structured, multi-phase qualitative coding process. This consisted of open and axial coding~\cite{charmaz2006constructing} for Terraform technical errors and OPA validation failures. The first author first performed the coding, and then the second and third authors reviewed the codes. Through an offline discussion, any disagreement about the codes was resolved.    

\paragraph{Step 1: Open Coding of Error Logs and Validation Failures}
In the first step, we conducted open coding on Terraform error logs. Guided by qualitative research principles~\cite{charmaz2006constructing}, we sought to answer the following questions: (1) What type of error does the log message describe? (2) What component of the Terraform configuration does the error affect? (3) Is the error syntactic, semantic, or environmental in nature?

Each error message was assigned a short, descriptive code that summarized its failure type (e.g., \texttt{missing variable}, \texttt{invalid attribute}, \texttt{resource type not found}). 
For example, the following error log was initially coded as \texttt{Invalid Argument}:

\begin{lstlisting}
Error: Unsupported argument

  on main.tf line 41, in resource "aws_route53_record" "weighted_record":
  41:     set_identifier = "replica-1"

An argument named "set_identifier" is not expected here.
\end{lstlisting}

In parallel, we also analyzed validation failures identified by custom OPA policies explicitly written to represent the intended outcomes of the IaC generation process. These policies were designed to verify that the generated infrastructure matches the desired configuration and resource usage. For each failed policy validation, the annotator manually reviewed the generated Terraform script against the original prompt to determine the cause of the intent violation. The analysis focused on identifying whether the generated infrastructure matched the user's specified requirements and, if not, characterizing the nature of the mismatch. Each failure was labeled with a descriptive code reflecting the underlying issue, such as \texttt{used deprecated resource}, \texttt{wrong resource type}, or \texttt{missing required resource}.

\paragraph{Step 2: Axial Coding into Broader Categories}
Building on the initial descriptive codes, we applied axial coding to organize related error types into broader conceptual categories based on the nature of the validation failure. This involved analyzing relationships between subcategories and identifying which validation rule was violated.

We organized related error codes by the type of validation they represented. Codes such as \texttt{missing variable}, \texttt{missing provider block}, and \texttt{missing module source} revealed patterns related to different aspects of Terraform validation.  These codes were categorized into syntax violations (basic Terraform language rules), schema mismatches (violations of provider specifications), runtime issues (execution-time problems), and intent misalignments (functional correctness). For instance, various argument-related errors, including \texttt{invalid argument}, \texttt{unsupported argument}, and \texttt{missing required argument}, were consolidated under the broader category of "Argument error" within the Schema validation category, as they all represented violations of provider schema specifications rather than basic syntax rules. 

\begin{figure}
    \centering
    \includegraphics[width=\linewidth]{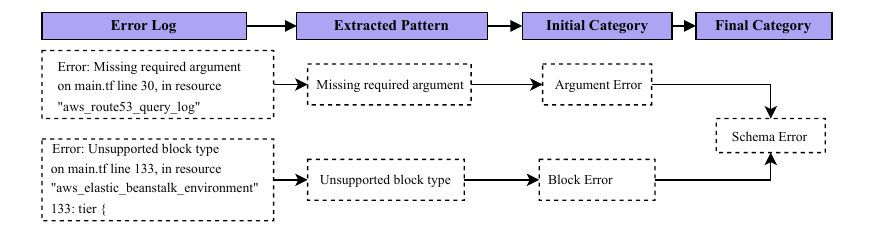}
    \caption{Example of How an Error is Classified into a Schema Error.}
    \label{fig:coding_example}
\end{figure}

Figure~\ref{fig:coding_example} illustrates this consolidation process, showing how multiple specific error instances are grouped into broader conceptual categories during axial coding. Individual error codes that initially captured specific manifestations were organized into higher-level error types that captured the underlying validation failure pattern.

\paragraph{Step 3: Development of the Second Dimension}
 
The second dimension of our taxonomy was developed by analyzing the underlying generation patterns that led to all observed failures. This dimension was informed by established LLM failure patterns in code generation research~\cite{tambon2025bugs, huang2025survey} and emerged from a systematic examination of the types of generation failure that caused each error manifestation across both technical and intent validation contexts.

Through analysis of the complete error corpus, we identified four primary types of generation failure: incompleteness, factual incorrectness, structural deficit, and contextual reasoning failure. For example, codes such as \texttt{missing variable} and \texttt{missing required resource} both reflected a pattern of \textit{incompleteness} in which the model omitted necessary elements. Conversely, \texttt{invalid argument} and \texttt{used deprecated resource} both reflected \textit{factual incorrectness} where the model generated elements that were invalid or outdated according to specifications. Similarly, \texttt{wrong resource type} indicated \textit{contextual reasoning failure} where the model failed to select appropriate resources for the given context. Finally, codes such as \texttt{format errors} and \texttt{Terraform language errors} are under \textit{structural deficit}, indicating that the generated code does not follow the language's syntax rules.

\subsubsection{Changelog Analysis}
Understanding whether LLM-generated code errors arise from knowledge staleness or hallucination is crucial~\cite{huang2025survey}. Errors due to invalid schema elements may indicate either that the elements were once valid but have since been deprecated (knowledge staleness), or they never existed and are fabricated content (hallucination)~\cite{tambon2025bugs}. This distinction helps determine if errors result from outdated training data or a deficit in domain-specific knowledge~\cite{huang2025survey, gao2023retrieval, pan2024unifying}. To investigate whether factual incorrectness errors were caused by knowledge staleness due to deprecated elements in the Terraform AWS provider, we analyzed provider changelogs from version 0.0.1 (June 2017) to version 5.96.0 (April 2025) \footnote{\url{https://github.com/hashicorp/terraform-provider-aws/blob/main/CHANGELOG.md}}. This involved extracting elements from error messages and cross-referencing them with historical deprecation records. 

The extraction and search rules were developed to identify relevant information from Terraform error messages and match them against changelog entries. These rules followed Terraform's documented changelog process \footnote{\url{https://github.com/hashicorp/terraform-provider-aws/blob/main/docs/changelog-process.md}}. This methodology facilitated the systematic classification of unsupported elements across the complete AWS provider changelog history, ensuring comprehensive coverage of deprecation events up to the model's estimated training cutoff.  Table~\ref{tab:extraction-rules} shows the used rules. 

\begin{table}[t]
\centering
\caption{Rules Defined for Each Extraction Type. [E] Rules are for Error Logs. [C] Rules are for the Changelog.}
\footnotesize
\begin{tabular}{|l|p{10.5cm}|}
\hline
\textbf{Rule Type} & \textbf{Rule} \\
\hline
[E] Argument Extraction & IF \texttt{msg} contains \texttt{"resource"} + \texttt{"<type>"} AND \texttt{"argument named"} + \texttt{"<arg>"}, THEN extract \texttt{type}, \texttt{arg}. \\
\hline
[E] Block Extraction & IF \texttt{msg} contains \texttt{"resource"} + \texttt{"<type>"} AND \texttt{"Blocks of type"} + \texttt{"<block>"}, THEN extract \texttt{type}, \texttt{block}. \\
\hline
[E] Resource Extraction & IF \texttt{msg} contains \texttt{"resource"} + \texttt{"<type>"} AND indicates unrecognized resource, THEN extract \texttt{type}. \\
\hline
[C] Date Context & IF line matches \texttt{"\#\# [version] (date)"}, THEN set \texttt{date\_context}. \\
\hline
[C] Entry Identification & IF line starts with \texttt{*} AND contains \texttt{target\_resource}, THEN examine entry. \\
\hline
[C] Element Matching & IF entry contains \texttt{"<element>"} with non-alphanumeric bounds, THEN match element. \\
\hline
[C] Deprecation Classification & IF entry contains \texttt{"deprecated"}, THEN classify as \texttt{deprecated}; ELSE classify as \texttt{never documented}. \\
\hline
\end{tabular}
\label{tab:extraction-rules}
\end{table}

\subsection{The NL2IaC Taxonomy}
In this section, we present our error taxonomy for LLM-generated Terraform code. The taxonomy classifies errors along two dimensions: the type of validation failure encountered (Dimension 1) and the underlying generation pattern that caused the error (Dimension 2). Because Terraform validation is phased, where technical errors block further functional evaluation, we present the taxonomy across two tables to provide appropriate context for the error counts. Table~\ref{tab:technical_errors_dim1} and Table~\ref{tab:technical_errors_dim2} detail technical validation errors, while Table~\ref{tab:intent_errors_dim1} and Table~\ref{tab:intent_errors_dim2} focus on intent validation errors identified only in scripts that passed technical validation.

\begin{table}[t]
\centering
\caption{Dimension 1: Technical Validation Errors in LLM-Generated Terraform Code}
\label{tab:technical_errors_dim1}
\setlength{\arrayrulewidth}{0.01mm}
\footnotesize
\renewcommand{\arraystretch}{1.2}
\begin{tabular}{|p{2.5cm}|p{2.8cm}|p{6cm}|r|}
\hline
\textbf{Error Category} & \textbf{Error Sub-Category} & \textbf{Atomic Error} & \textbf{GPT-4o} \\
\hline

\multirow{2}{*}{Syntax}
  & Terraform language error & Breaks basic syntax rules or structure & \cellcolor{color2}4 \\ 
\cline{2-4}
  & Format error & Uses invalid names or formatting & \cellcolor{color2}8 \\ 
\hline

\multirow{16}{*}{Schema}
  & \multirow{6}{*}{Argument error}
    & Uses an argument that is not allowed & \cellcolor{color1}326 \\ 
  &   & Leaves out a required argument & \cellcolor{color4}140 \\ 
  &   & Assigns a wrong or unsupported value & \cellcolor{color1}13 \\ 
  &   & Uses a reserved word as an argument name & \cellcolor{color1}2 \\ 
  &   & Combines arguments that conflict & \cellcolor{color3}3 \\ 
  &   & Repeats the same argument twice & \cellcolor{color3}4 \\ 
\cline{2-4}
  & \multirow{2}{*}{Attribute error}
    & Refers to an attribute that does not exist & \cellcolor{color1}3 \\ 
  &   & Uses an attribute in the wrong way & \cellcolor{color3}5 \\ 
\cline{2-4}
  & \multirow{3}{*}{Block error}
    & Adds a block type that is not supported & \cellcolor{color1}104 \\ 
  &   & Misses blocks that are required & \cellcolor{color4}64 \\ 
  &   & Adds more blocks than allowed & \cellcolor{color3}1 \\ 
\cline{2-4}
  & \multirow{3}{*}{Resource error}
    & Uses a resource type that does not exist & \cellcolor{color1}44 \\ 
  &   & Refers to a resource that is invalid or missing & \cellcolor{color1}10 \\ 
  &   & Declares the same resource more than once & \cellcolor{color3}12 \\ 
\hline
\multirow{4}{*}{Runtime}
  & \raggedright Environment error & Refers to a file that cannot be found & \cellcolor{color3}23 \\ 
\cline{2-4}
  & \multirow{3}{*}{Provider error}
    & Refers to a provider that is not configured & \cellcolor{color4}1 \\ 
  &   & Configures the same provider multiple times & \cellcolor{color3}6 \\ 
\cline{2-4}
  & Version error & Uses an invalid or unsupported version range & \cellcolor{color1}1 \\ 
\hline

\end{tabular}
\end{table}

\begin{table}
\centering
\caption{Dimension 2: Technical Validation Errors in LLM-Generated Terraform Code}
\label{tab:technical_errors_dim2}
\setlength{\arrayrulewidth}{0.01mm}
\footnotesize
\begin{tabular}{|c|c|c|}
\hline
\textbf{Error Category} & \textbf{Error Sub-Category} & \textbf{GPT-4o} \\
\hline
\multirow{4}{*}{\shortstack[c]{LLM Generation\\Failure Type}} 
  &  Factual Incorrectness & \cellcolor{color1}503 \\ 
  & Structural Deficit & \cellcolor{color2}12 \\ 
  & Contextual Reasoning Failure & \cellcolor{color3}54 \\ 
  & Incompleteness & \cellcolor{color4}205 \\
\hline
\end{tabular}
\end{table}

\begin{table}[t]
\centering
\caption{Dimension 1: Intent Validation Errors in LLM-Generated Terraform Code}
\label{tab:intent_errors_dim1}
\setlength{\arrayrulewidth}{0.01mm} 
\begingroup
\footnotesize
\renewcommand{\arraystretch}{1.2}
\begin{tabular}{|p{2.5cm}|p{2.8cm}|p{6cm}|r|}
\hline
\textbf{Error Category} & \textbf{Error Sub-Category} & \textbf{Atomic Error} & \textbf{GPT-4o} \\
\hline
\multirow{4}{*}{Intent}
  & \multirow{3}{*}{Misaligned resource}
    & Fails to include a necessary resource & \cellcolor{color4}12 \\ 
  &   & Uses a resource that does not match the intent & \cellcolor{color3}14 \\ 
  &   & Uses a resource that is outdated or deprecated & \cellcolor{color1}10 \\ 
\cline{2-4}
  & Misconfiguration & Uses correct resources but misconfigured them & \cellcolor{color3}6 \\ 
\hline 
\end{tabular}
\endgroup
\end{table}

\begin{table}[t]
\centering
\centering
\caption{Dimension 2: Intent Validation Errors in LLM-Generated Terraform Code}
\label{tab:intent_errors_dim2}
\setlength{\arrayrulewidth}{0.01mm}
\footnotesize
\begin{tabular}{|c|c|c|}
\hline
\textbf{Error Category} & \textbf{Error Sub-Category} & \textbf{GPT-4o} \\
\hline
\multirow{4}{*}{\shortstack[c]{LLM Generation\\Failure Type}} %
  & Factual Incorrectness & \cellcolor{color1}10 \\ 
  & Structural Deficit & \cellcolor{color2}0 \\   
  & Contextual Reasoning Failure & \cellcolor{color3}20 \\ 
  & Incompleteness & \cellcolor{color4}12 \\ 
\hline
\end{tabular}
\end{table}

\subsubsection*{Dimension 1: Error Manifestation Category}
This dimension, presented in Tables~\ref{tab:technical_errors_dim1} and~\ref{tab:intent_errors_dim1}, classifies errors by the validation context in which they are identified: either during Terraform's technical validation or during subsequent intent validation.

The first three categories represent errors identified during technical validation. \textbf{Syntax errors} are violations of Terraform language rules that prevent successful parsing. \textbf{Schema errors} are violations against the structure, arguments, or types defined by Terraform core or providers. These signify non-conformance with the specific schema contract, such as using unrecognized arguments, omitting mandatory ones, or assigning incorrect data types. \textbf{Runtime errors} are due to issues like API interaction problems, conflicts with existing infrastructure, missing environment dependencies, or version incompatibilities encountered during execution.

The final category represents failures identified during intent validation, occurring after technical validation has passed. \textbf{Intent errors} occur when the generated infrastructure, even if technically valid, fails intent validation. This means the resulting infrastructure does not meet the user's specified requirements derived from the prompt.

\subsubsection*{Dimension 2: LLM Generation Failure Type}
 
Dimension 2, reported in Table~\ref{tab:technical_errors_dim2} and~\ref{tab:intent_errors_dim2}, classifies the type of failure inferred to have occurred within the LLM during code generation. The related work on NL-to-Code generation for tasks such as SQL or general-purpose programming often organizes errors along syntactic and semantic dimensions. This is facilitated by a relatively direct mapping between the natural language prompt and the generated code~\cite{ning2024insights, wang2024large}. In NL2IaC, however, the task involves generating a detailed, structured configuration from a high-level description of a desired system state. This mapping is less direct and requires the LLM model to recall deeper knowledge, engage in contextual reasoning, and plan at the structural level. As a result, errors observed in NL2IaC reflect broader generation failures rather than localized syntactic or semantic misunderstandings. Therefore, instead of relying solely on semantic classifications, the second dimension characterizes the types of generation failures inferred from the code itself. The specific categories adopted are informed by observed LLM failure patterns in code generation~\cite{tambon2025bugs, huang2025survey}.

The first category, \textbf{Factual Incorrectness}, applies when the generated code contains elements (e.g., resource types, arguments, values) that are invalid, nonexistent, deprecated, or incompatible with Terraform specifications or provider schemas. This indicates that the output does not conform to documented facts or requirements. The second category, \textbf{Structural Deficit}, covers failures to produce code that conforms to Terraform's syntax and structure. It includes syntax errors, incorrect block nesting, and structural formatting issues, indicating difficulties with the language's grammar and form. The third category, \textbf{Contextual Reasoning Failure}, involves failures related to context, consistency, or constraints resulting in incorrectly generated code. This category includes inconsistencies within the generated configuration, such as conflicting arguments within a resource or duplicate declarations. It also covers mismatches between the generated code and the inferred requirements or context from the user prompt, for example, selecting an inappropriate resource type or configuration for the stated goal.
The final category, \textbf{Incompleteness}, applies when the LLM omits elements essential for correctness or function, according to Terraform's schema or inferred requirements. This includes missing mandatory arguments, required nested blocks, or entire Terraform resources implicitly needed to fulfill the request.

\subsection{Error analysis}
To analyze the prevalence and nature of errors generated by the LLM, we leverage our two-dimensional taxonomy to explore the first question:
\textbf{RQ1}: \textbf{What are the characteristics and possible causes of errors made by Large Language Models when generating Infrastructure as Code in Terraform?}

\subsubsection{Error Types and Distribution} \label{subsubsec:error_distribution} 
To understand the types and distribution of errors in LLM-generated Terraform scripts, we reported the cross-distribution of error categories (Dimension 1) and LLM generation failure types (Dimension 2), as presented in Table~\ref{tab:phase1_2d_dist_percent} for technical validation errors and Table~\ref{tab:intent_2d_dist_percent} for intent validation errors. 

Schema-related issues overwhelmingly dominate, constituting 94.5\% of all technical validation failures. Within this category, Argument errors are the most prevalent, accounting for 63.1\% of technical errors. Considering LLM generation failure types (Dimension 2), Factual Incorrectness emerges as the leading cause of technical validation failures, responsible for 65.0\% of instances, followed by incompleteness accounting for 26.5\% of technical errors.

Contextual Reasoning Failure, responsible for 47.6\% of all instances, is the most frequent LLM generation failure type associated with intent errors. The most common specific error is the selection of a resource that does not meet the user's intent (33.3\%, linked to Contextual Reasoning Failure) and the omission of a necessary resource (28.6\%, associated with Incompleteness).

\begin{table}[t]
\centering
\caption{Cross-Distribution of Technical Error Types (Dim 1) vs. Error Patterns (Dim 2) as Percentage of Total Technical Instances (\%)}
\label{tab:phase1_2d_dist_percent}
\sisetup{round-mode=places, round-precision=1, table-format=2.1}
\footnotesize
\begin{tabular}{@{}l S[table-format=2.1] S[table-format=1.1] S[table-format=1.1] S[table-format=2.1] S[table-format=2.1]@{} | S[table-format=2.1]@{}}
\toprule
\textbf{Error Category / Type} & {\textbf{\FI}} & {\textbf{\SD}} & {\textbf{\CRF}} & {\textbf{\Incomp}} & {\textbf{Row Total}} & {\textbf{Category Raw Total}} \\
\textbf{(Dimension 1)}         & {(\%)}         & {(\%)}         & {(\%)}          & {(\%)}           & {(\%)}                      & {(\%)}\\
\midrule
\textit{Syntax} & & & & & & \\
\quad Terraform  language error & 0.0 & 0.5 & 0.0 & 0.0 & 0.5 & 1.5\\
\quad Format error & 0.0 & 1.0 & 0.0 & 0.0 & 1.0 & \\
\midrule
\textit{Schema} & & & & & & \\
\quad Argument error & 44.1 & 0.0 & 0.9 & 18.1 & 63.1 & \\
\quad Attribute error & 0.4 & 0.0 & 0.6 & 0.0 & 1.0 & 94.5\\
\quad Block error & 13.4 & 0.0 & 0.1 & 8.3 & 21.8 & \\
\quad Resource error & 7.0 & 0.0 & 1.6 & 0.0 & 8.6 & \\
\midrule
\textit{Runtime} & & & & & & \\
\quad Environment error & 0.0 & 0.0 & 3.0 & 0.0 & 3.0 & \\
\quad Provider error & 0.0 & 0.0 & 0.8 & 0.1 & 0.9 & 4.0\\
\quad Version error & 0.1 & 0.0 & 0.0 & 0.0 & 0.1 & \\
\midrule
\textbf{Column Total (\%)} & \textbf{65.0} & \textbf{1.5} & \textbf{7.0} & \textbf{26.5} & \textbf{100.0} \\
\bottomrule
\end{tabular}
\end{table}

\begin{table}[t]
\centering
\caption{Cross-Distribution of Intent Error Types (Dim 1) vs. Error Patterns (Dim 2) as Percentage of Total Intent Instances (\%)}
\label{tab:intent_2d_dist_percent}
\sisetup{round-mode=places, round-precision=1, table-format=2.1}
\footnotesize
\begin{tabular}{@{} l S[table-format=2.1] S[table-format=1.1] S[table-format=2.1] S[table-format=2.1] S[table-format=3.1] @{}}
\toprule
\textbf{Error Type} & {\textbf{\FI}} & {\textbf{\SD}} & {\textbf{\CRF}} & {\textbf{\Incomp}} & {\textbf{Row Total}} \\
\textbf{(Dimension 1 - Intent)} & {(\%)}         & {(\%)}         & {(\%)}          & {(\%)}           & {(\%)}           \\
\midrule
Fails to include necessary resource & 0.0 & 0.0 & 0.0 & 28.6 & 28.6 \\
Uses resource not matching intent & 0.0 & 0.0 & 33.3 & 0.0 & 33.3 \\
Uses outdated/deprecated resource & 23.8 & 0.0 & 0.0 & 0.0 & 23.8 \\
Misconfiguration                  & 0.0 & 0.0 & 14.3 & 0.0 & 14.3 \\
\midrule
\textbf{Column Total (\%)} & \textbf{23.8} & \textbf{0.0} & \textbf{47.6} & \textbf{28.6} & \textbf{100.0} \\
\bottomrule
\end{tabular}
\end{table}

\subsubsection{Analysis of Underlying Causes for LLM Generation Errors}

We leverage the previously identified error patterns. Notably, Schema errors (driven by Factual Incorrectness and Incompleteness) dominate technical validation, while Contextual Reasoning Failure is prevalent in intent validation. Our analysis links the primary LLM Generation Failure patterns (Factual Incorrectness, Incompleteness, and Contextual Reasoning Failure) to potential underlying causes: Knowledge Gaps, Model Limitations, and External Changes. This analysis integrates quantitative results, insights from your changelog analysis, and qualitative interpretation. 

\paragraph{Factual Incorrectness: Hallucination vs. Knowledge Staleness}
Factual Incorrectness errors are highly prevalent during technical validation (65.0\%). They manifest as generating elements that are invalid according to the Terraform specification or provider schemas. To better understand their causes, we examined the three most frequent types during technical validation: unsupported arguments, unsupported blocks, and unsupported resources (as shown in Table~\ref{tab:hallucination_distribution}).

In Terraform, deprecation involves a warning period before removal. An element's removal only causes a validation error after the next major provider release\footnote{\url{https://developer.hashicorp.com/terraform/plugin/framework/deprecations}}. Therefore, deprecated elements continue to pass technical validation until they are explicitly removed. To assess whether the technical errors stemmed from using elements removed before the model’s likely training cutoff (gpt-4o, October 2023), we compared the errors against the AWS provider changelog. We noted that the last major release occurred in May 2023.

\begin{table}[t]
    \footnotesize
    \centering
    \caption{Distribution of Unsupported Arguments, Blocks, and Resources in Technical Validation Errors}
    \label{tab:hallucination_distribution}
    \begin{tabular}{c|cc|cc|cc}
     & \multicolumn{2}{l|}{Unsupported Argument} & \multicolumn{2}{l|}{Unsupported Block} & \multicolumn{2}{l}{Unsupported Resources} \\ \cline{2-7} 
     & Count & \% & Count & \% & Count & \% \\ \hline
    Hallucinated & 309 & 94.8 & 102 & 98.1 & 44 & 100 \\ \hline
    Deprecated & 17 & 5.2 & 2 & 1.9 & 0 & 0
    \end{tabular}

\end{table}

The analysis reveals that only a small fraction of unsupported argument errors (5.2\%) and unsupported block errors (1.9\%) involved elements deprecated and potentially removed before the cutoff (see Table~\ref{tab:hallucination_distribution}). No unsupported resource errors corresponded to previously valid resources. This suggests that the model maintained relatively accurate knowledge regarding provider schema removals up to its training cutoff date.

However, the majority of these technical Factual Incorrectness errors stem from hallucinated schema elements—arguments, blocks, or resources that seemingly never existed according to any recorded Terraform specification. Despite possessing up-to-date knowledge of removals, the model frequently struggles to accurately recall schemas when generating configurations. This points towards a Model Limitation in reliably accessing or generating known facts, rather than simply reflecting outdated knowledge~\cite{huang2025survey, tambon2025bugs}.

A likely cause for these hallucinated schema elements is the nature of the model's training data. Rather than being trained directly on authoritative provider schemas, the model likely encountered partial, example-driven representations of infrastructure code across various public sources~\cite{zhong2024can}. In particular, publicly available Terraform code often involves significant abstraction through user-defined modules, variable templates, and project-specific naming conventions. This real-world variability can introduce arguments and structural patterns that do not exist in the underlying provider specifications. Research on IaC code quality has identified various anti-patterns and smells in real-world IaC repositories~\cite{schwarz2018code,rahman2020code,bessghaier2024prevalence}, further highlighting the potential for training data to contain suboptimal or non-standard patterns. As a result, the model may overgeneralize common configurations across resources or absorb invalid patterns from noisy training examples. LLMs can struggle with "long-tail knowledge" (less frequent information) and may exhibit "deficits in domain-specific knowledge," sometimes leading to "factual fabrication"~\cite{huang2025survey, gao2023retrieval, pan2024unifying}.

Furthermore, LLMs are optimized to produce linguistically plausible outputs rather than to verify strict schema compliance~\cite{brown2020language}. Together, these factors can contribute to the generation of arguments, blocks, or resources that appear plausible but do not correspond to any actual Terraform specification.

A different pattern emerges during intent validation. Here, Factual Incorrectness manifests as the use of outdated or deprecated resources (23.8\% of intent errors). These resources pass technical validation but fail functional requirements. This suggests that External Changes, such as ongoing evolution and deprecation post-cutoff, significantly impact functional correctness~\cite{tambon2025bugs}. This occurs even if technical validity is maintained initially due to the deprecation warning period. Therefore, while technical FI errors are mainly driven by hallucination (Model Limitation), intent-level FI errors appear more susceptible to Knowledge Gaps caused by knowledge staleness.

\paragraph{Incompleteness}, which accounts for a significant share of both technical (26.5\%) and intent (28.6\%) errors, arises when the LLM fails to include necessary configuration elements. A further analysis was conducted to determine whether these errors, particularly during technical validation, could be attributed to knowledge staleness. Specifically, the introduction of newly required arguments or blocks after the model's training cutoff. But comparing the missing elements against the AWS provider changelog revealed that only two instances corresponded to schema changes.

This strongly indicates that knowledge staleness is not a significant driver for technical Incompleteness errors; rather, these omissions primarily arise from internal model limitations in sequential generation and multi-step reasoning~\cite{wei2022chain, brown2020language}.

Additional analysis of missing elements revealed patterns in those technical argument omissions:
\textbf{Simple identifiers} such as \texttt{name}, \texttt{user\_name}, and \texttt{pipeline\_name} are often omitted, suggesting issues with basic attribute generation or focus during the generation process.
\textbf{Cross-resource references}, including \texttt{vpc\_id}, \texttt{subnet\_ids}, and \texttt{target\_vault\_name}, indicate failures in modeling resource interdependencies. The model struggles to dynamically link argument values to other resources' attributes, revealing limitations in its dependency tracking.
\textbf{Low-level configuration settings} like \texttt{enabled}, \texttt{status}, and \texttt{rule\_action} are inconsistently handled, reflecting challenges with fine-grained configurations.

Similarly, the omission of required nested blocks often involves complex structures. These typically include constrained elements, such as those needing a specific configuration type from multiple options. The failure to generate these intricate, structurally complex blocks highlights the model's limitations in handling detailed structural requirements and the logical constraints inherent in many IaC schemas.

Incompleteness also manifests during intent validation, though in a different form. While technical validation incompleteness involves missing required arguments and blocks based on strict schema rules, intent validation reveals cases where the generated infrastructure configuration lacks entire resources necessary to fulfill the user's intended goal. This often occurs when required resources are not explicitly named in the prompt and must be inferred from implicit domain knowledge, representing a different class of model limitation related to contextual understanding rather than schema recall.

\paragraph{Contextual Reasoning Failures} during intent validation (47.6\% of intent errors) primarily manifest in two ways: \textit{Misaligned Resource} selection and \textit{Resource Misconfiguration}.

Misaligned Resource errors occur when the model selects a technically valid resource that fails to meet the user's specified intended goal. For instance, when prompted to create a \textit{serverless} MSK cluster:

\begin{tcolorbox}[colback=gray!5, colframe=gray!80, title=Prompt \#210]
Create a serverless MSK cluster with 3 broker nodes in us-east-1.
\end{tcolorbox}

Instead of the correct \texttt{aws\_msk\_serverless\_cluster}, the model generated an \texttt{aws\_msk\_cluster} resource. This suggests difficulty with fine-grained semantic disambiguation between similar resource types, even when explicitly prompted. Such errors may stem from overgeneralization, where the model defaults to a more common resource type when faced with closely related alternatives. \\

Misconfiguration errors, conversely, involve selecting the correct resource type but configuring it improperly for the intended purpose.
In another example, the model correctly identified the need for an Elastic Beanstalk environment but missed a critical setting from the prompt:

\begin{tcolorbox}[colback=gray!5, colframe=gray!80, title=Prompt \#27]
Create an Elastic Beanstalk WebServer environment with an Elastic Beanstalk application. Name the IAM role eb\_ec2\_role, the instance profile eb\_ec2\_profile, and the application my\_application.
\end{tcolorbox}

While correctly creating the \texttt{aws\_elastic\_beanstalk\_environment} and related components with appropriate names, the model failed to set the required \texttt{tier="WebServer"}. This type of omission might occur because specific configuration attributes, such as environment tiers, appear less frequently in the training data than basic resource-creation patterns. Consequently, the model may overlook such settings despite explicit instructions.

\subsubsection{Summary of the Answers to RQ1}
We systematically developed a taxonomy of errors encountered when generating Terraform scripts with LLMs. It consists of 19 error types. The analysis of those errors in the IaC-Eval benchmark dataset shows that Schema Errors dominate technical validation failures (94.5\%), with Factual Incorrectness (65.0\%) and Incompleteness (26.5\%) as the primary generation failure patterns. Intent validation errors show a different distribution, with Contextual Reasoning Failure being most prevalent (47.6\%). The findings reveal that while basic Terraform syntax generation is relatively robust, significant challenges impede the reliable production of correct, intent-aligned IaC.

Considering the underlying causes of the LLM failures, our analysis distinguished between different failure mechanisms across validation contexts. Technical validation failures are dominated by Schema errors, primarily Factual Incorrectness (invalid arguments, blocks, resources) and Incompleteness (missing required arguments, blocks). Our analysis, supported by cross-referencing with provider changelogs, indicates that technical Factual Incorrectness often stems from model hallucination of non-existent schema elements, rather than solely from knowledge staleness regarding recently removed components~\cite{tambon2025bugs}. This points to a deficit in domain-specific knowledge, a known limitation where models struggle to memorize less frequent information~\cite{pan2024unifying, gao2023retrieval}. Technical Incompleteness appears primarily driven by model limitations in planning and handling complex structural requirements~\cite{tambon2025bugs}, not by outdated knowledge of requirements. 
Conversely, errors identified during intent validation present different challenges. Contextual Reasoning Failures (resource misalignment, misconfiguration) and the use of outdated or deprecated elements (a form of Factual Incorrectness) point towards the impact of knowledge staleness post-training, difficulties in fine-grained semantic disambiguation, limitations in maintaining logical consistency, and failures in accurately translating user intent into correctly configured resources.

\section{Configuration Knowledge Injection for IaC Code Generation}\label{sec:methods}

The preceding section established a clear understanding of the challenges facing LLM-based IaC generation and identified specific failure patterns that limit current approaches. The prevalence of errors rooted in inaccurate, hallucinated, or missing schema knowledge (Factual Incorrectness and Incompleteness) strongly suggests a need for better knowledge grounding. We hypothesize that generation accuracy can be substantially improved by providing the LLM with explicit access to structured and up-to-date configuration knowledge, consisting of provider schemas, dependencies, and documentation. The significant challenges observed in contextual reasoning failures, handling complex dependencies, and accurately applying configuration details further support the need for enhanced knowledge access during generation. In this section, by designing and implementing multiple knowledge representation and retrieval strategies, we investigate how structured domain knowledge can improve both the technical correctness and the alignment of intent in generated Terraform code. The primary research question is \textbf{RQ2:} \textbf{How can configuration knowledge improve the overall quality of LLM-generated Terraform code?}

\subsection{System Architecture of IaC Generator}  
\label{sec:method_system_architecture}
This section presents the architecture of the IaC generation system, developed to investigate how injecting configuration knowledge improves LLM-generated Terraform code. The system implements two Retrieval-Augmented Generation (RAG) approaches, Naive RAG~\cite{fan2024survey,siriwardhana2023improving} and Graph RAG~\cite{edge2024local,he2024g}, within a unified framework. We will rationalize the use of Graph RAG and its enhancements later when we describe them in detail.  

\begin{figure}[t]
    \centering
    \includegraphics[width=0.8\linewidth]{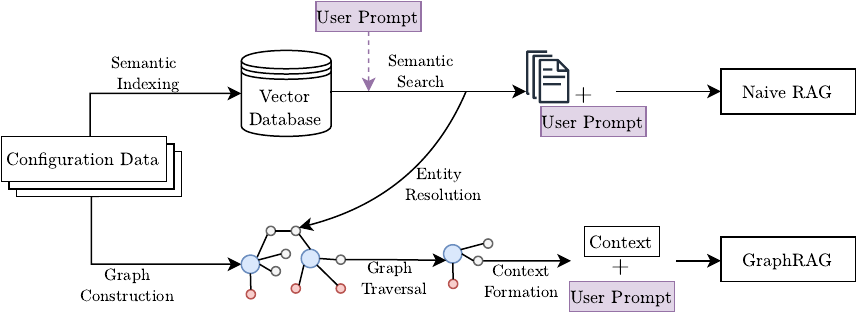}
    \caption{High-Level System Architecture of IaC Generator}
    \label{fig:system_architecture}
\end{figure}

Figure~\ref{fig:system_architecture} illustrates the high-level system architecture. The system operates in two phases: offline knowledge base construction and online query processing.
Configuration data for the Terraform providers undergoes two parallel processing paths. Semantic Indexing creates vector embeddings stored in a vector database, while Graph Construction constructs a knowledge graph that encodes resources, arguments, and relationships. The vector database serves both RAG approaches, while the knowledge graph is exclusive to Graph RAG.

Both systems begin with Semantic Search, where user queries are embedded and matched against the vector database to retrieve relevant documentation chunks. In Naive RAG, these chunks are directly combined with the user prompt for generation. In Graph RAG, semantic search results undergo Entity Resolution to identify corresponding knowledge graph nodes. Graph Traversal then leverages knowledge of the graph structure to retrieve comprehensive configuration information (arguments, blocks, and relationships), which is formatted through Context Formation before generation.

This modular architecture enables fair comparison between approaches while supporting incremental enhancements such as improved semantic matching, reference relationship extraction, and iterative repair mechanisms. The following sections discuss the implementation of each component in detail. Please note that, as the design of prompts is not a primary focus of this study, we included the prompts used by the LLM in our RAG systems in our replication package. 

\subsection{Knowledge Sources and Preparation}
\label{sec:knowledge_sources}
\begin{figure}[t]
  \centering
  \includegraphics[width=0.7\textwidth]{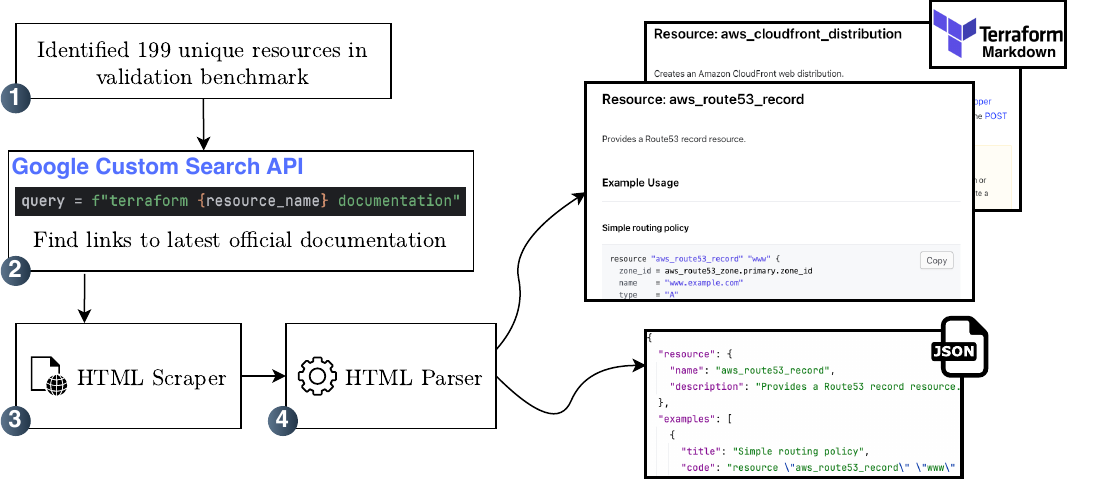}
  \caption{Pipeline for Data Acquisition and Initial Processing. Starting from identified resources (1), the system queries for official documentation URLs (2), scrapes HTML content (3), and parses it into structured formats (4). The outputs include Markdown for textual content and JSON for structured data extraction.}
  \label{fig:knowledge_scraping}
\end{figure}

The configuration knowledge for this research was derived from the official Terraform documentation for the AWS providers, as the IaC-Eval benchmark is only for AWS. We used the Terraform AWS provider documentation version 5.90.0\footnote{\url{https://registry.terraform.io/providers/hashicorp/aws/5.90.0/docs}}  as the primary source of configuration knowledge. The documentation follows a standardized template that includes resource descriptions, argument references, attribute references, and usage examples. Figure~\ref{fig:knowledge_scraping} illustrates the automated pipeline developed to acquire and process Terraform documentation. Starting with 199 unique Terraform resources identified in the validation benchmark, the Google Custom Search API locates each corresponding official documentation page using queries formatted as \texttt{"terraform \{resource\_name\} documentation"}. HTML content is then scraped and parsed into Markdown files for human readability and JSON for systematic processing. 
The acquired documentation underwent several processing steps to prepare it for knowledge base construction. Markdown files served as a corpus for semantic indexing, while structured JSON representations were used for schema extraction in the knowledge graph. Despite standardized templates, formatting inconsistencies hindered the reliability of schema extraction. To overcome this, we employed \textit{tfschema}\footnote{\url{https://github.com/minamijoyo/tfschema}}, which extracts provider schemas from compiled Terraform binaries, providing accurate structural definitions. 
The final dataset combines human-authored documentation for semantic richness and machine-extracted schemas for structural accuracy, forming the foundation for both vector-based embeddings and structured knowledge graph representations. 

\subsection{Knowledge Representation}
\label{sec:knowledge_representation}
To effectively augment LLM with configuration knowledge, the processed Terraform documentation must be transformed into machine-readable representations. This section describes two complementary knowledge representation strategies: semantic embeddings stored in a vector database for similarity-based retrieval, and a knowledge graph encoding explicit relationships between configuration elements. Both representations derive from the same source documentation and serve as foundations for the Naive RAG and Graph RAG systems, respectively.

\subsubsection{Semantic Embeddings}
\label{ssec:sem-emb}
\begin{figure}[t]
  \centering
  \includegraphics[width=\textwidth]{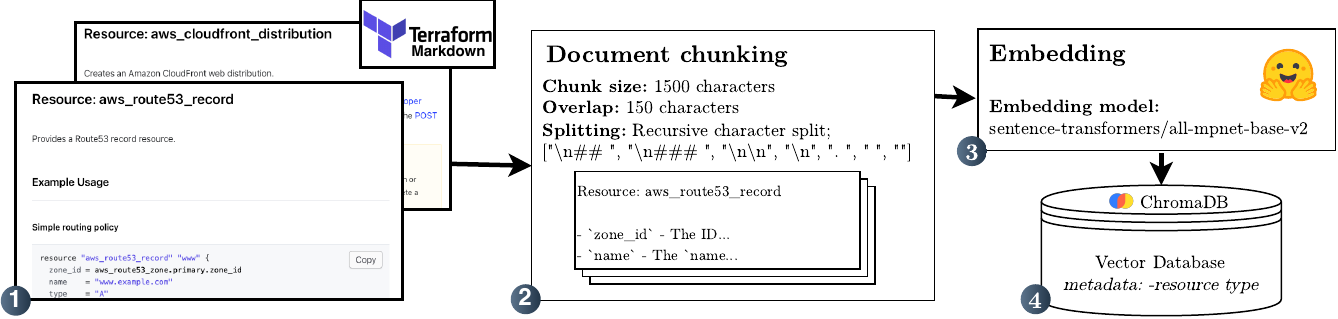} 
  \caption{Semantic Embedding Pipeline. Raw Terraform Markdown documentation (1) is segmented into overlapping chunks using recursive character splitting (2). Each chunk is transformed into a 768-dimensional vector using sentence-transformers (3), then stored with resource metadata in ChromaDB (4).}
  \label{fig:semantic_embeddings}
\end{figure}

The semantic indexing approach transforms Terraform documentation into high-dimensional vector representations to capture conceptual meaning and enable similarity-based retrieval~\cite{reimers2019sentence, karpukhin2020dense}. This representation serves as the foundational knowledge source for the Naive RAG baseline, ensuring reliable performance comparisons with graph-based methods. Figure~\ref{fig:semantic_embeddings} illustrates the complete semantic indexing pipeline, from raw Terraform documentation through chunking, embedding generation, and storage in the vector database. 

The Terraform documentation corpus is segmented into discrete chunks using recursive character-based splitting, with a chunk size of 1,500 characters and a 150-character overlap. These parameters balance semantic coherence with computational constraints, preserving complete argument descriptions and code examples within single chunks. 

Recursive splitting prioritizes section headers (\texttt{\textbackslash n\#\# } and \texttt{\textbackslash n\#\#\# }), paragraph boundaries (\texttt{\textbackslash n\textbackslash n}), line breaks (\texttt{\textbackslash n}), sentence endings (\texttt{. }), and finally individual characters when necessary. The 150-character overlap between chunks ensures that important configuration details spanning chunk boundaries maintain contextual integrity.

Vector embeddings are generated using the \texttt{sentence-transformers/all-mpnet-base-v2} model \footnote{\url{https://huggingface.co/sentence-transformers/all-mpnet-base-v2}}, which represents the current state-of-the-art for general-purpose sentence embedding~\cite{reimers2019sentence}. Each documentation chunk is converted into a 768-dimensional dense vector representation, facilitating efficient retrieval of contextually relevant segments during query processing via cosine similarity measures.

The generated embeddings are indexed and stored in ChromaDB~\footnote{\url{https://www.trychroma.com/}}, a vector database optimized for similarity search. Each embedding includes metadata indicating the source resource type, enabling both semantic-similarity retrieval and resource-specific filtering.

\subsubsection{Knowledge Graph}
\label{ssec:knowledge_graph}
While semantic embeddings capture textual similarity, they cannot explicitly represent the structured relationships needed for valid Terraform configurations. The knowledge graph representation addresses this limitation by encoding Terraform's domain structure as a directed, labeled graph where entities and their relationships are explicitly modeled~\cite{hogan2021knowledge}.

\begin{figure}[t]
    \centering
    \begin{subfigure}[b]{0.45\textwidth}
        \includegraphics[width=\textwidth]{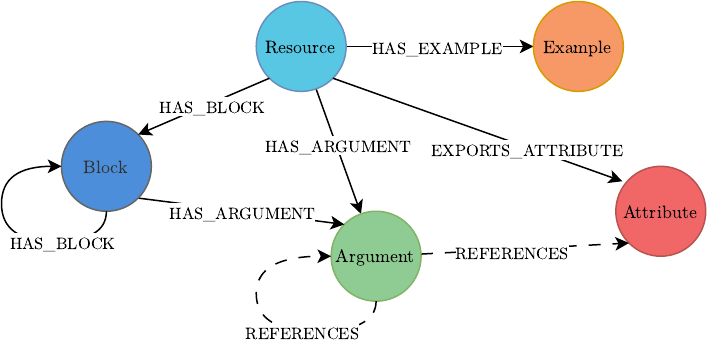} 
        \caption{Core Schema with Entity Types and Base Relationships.}
        \label{fig:kg_schema}
    \end{subfigure}
    \hfill
    \begin{subfigure}[b]{0.45\textwidth}
        \includegraphics[width=\textwidth]{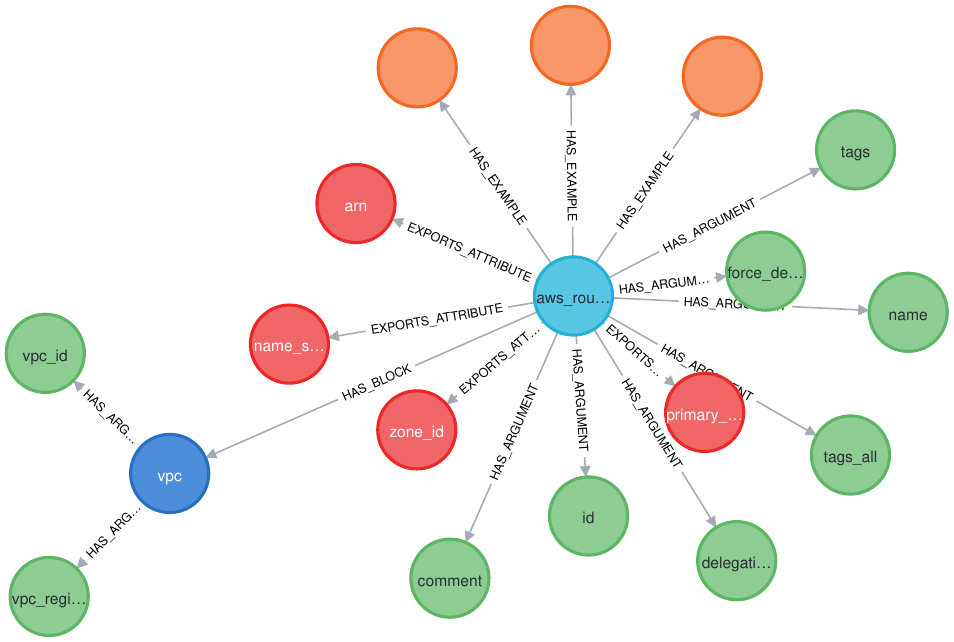} 
        \caption{Example Instantiation Showing \texttt{aws\_route53\_zone}.}
        \label{fig:kg_example}
    \end{subfigure}
    \caption{Knowledge Graph Schema Design and Example. Resources (blue), Arguments (green), and Attributes (red) are connected through typed relationships, enabling structured retrieval.}
    \label{fig:kg_overview}
\end{figure}

The knowledge graph schema captures Terraform's hierarchical configuration model through five entity types (Table~\ref{tab:kg_entities}) and their relationships (Table~\ref{tab:kg_relationships}). Figure~\ref{fig:kg_schema} presents the schema design, while Figure~\ref{fig:kg_example} demonstrates its application to the \texttt{aws\_route53\_zone} resource.

\begin{table}[t]
\centering
\caption{Knowledge Graph Entity Types and Their Properties}
\footnotesize
\label{tab:kg_entities}
\begin{tabular}{p{2cm}p{6cm}p{5cm}}
\toprule
\textbf{Entity Type} & \textbf{Description} & \textbf{Properties} \\
\midrule
\multirow{2}{2cm}{\texttt{Resource}} & \multirow{2}{6cm}{Terraform resource types representing cloud infrastructure components} & \texttt{name}: Resource identifier\\
& & \texttt{description}: Purpose and usage\\
\addlinespace
\hline
\addlinespace
\multirow{6}{2cm}{\texttt{Argument}} & \multirow{6}{6cm}{Configurable parameters controlling resource behavior} & \texttt{name}: Argument identifier\\
& & \texttt{description}: Parameter purpose\\
& & \texttt{type}: Data type specification\\
& & \texttt{required}: Boolean constraint\\
& & \texttt{id}: Hierarchical path\\
& & \texttt{resource}: Parent resource\\
\addlinespace
\hline
\addlinespace
\multirow{5}{2cm}{\texttt{Block}} & \multirow{5}{6cm}{Nested configuration structures grouping related arguments} & \texttt{name}: Block identifier\\
& & \texttt{description}: Block purpose\\
& & \texttt{cardinality}: Min/max occurrences\\
& & \texttt{id}: Hierarchical path\\
& & \texttt{resource}: Parent resource\\
\addlinespace
\hline
\addlinespace
\multirow{4}{2cm}{\texttt{Attribute}} & \multirow{4}{6cm}{Read-only properties exposed after resource creation} & \texttt{name}: Attribute identifier\\
& & \texttt{description}: Output meaning\\
& & \texttt{type}: Data type\\
& & \texttt{resource}: Parent resource\\
\addlinespace
\hline
\addlinespace
\multirow{4}{2cm}{\texttt{Example}} & \multirow{4}{6cm}{Code snippets demonstrating resource usage} & \texttt{name}: Example title\\
& & \texttt{code}: Terraform code\\
& & \texttt{index}: Example number\\
& & \texttt{resource}: Parent resource\\
\bottomrule
\end{tabular}
\end{table}

\begin{table}[t]
\centering
\caption{Knowledge Graph Relationship Types}
\label{tab:kg_relationships}
\footnotesize
\begin{tabular}{p{2.5cm}p{3.5cm}p{7cm}}
\toprule
\textbf{Relationship} & \textbf{Source → Target} & \textbf{Description} \\
\midrule
\texttt{HAS\_ARGUMENT} & Resource → Argument & Links resources to their configurable parameters\\
& Block → Argument & Links blocks to their nested arguments\\
\addlinespace
\hline
\addlinespace
\texttt{HAS\_BLOCK} & Resource → Block & Connects resources to nested configuration structures\\
& Block → Block & Enables recursive nesting of configuration blocks\\
\addlinespace
\hline
\addlinespace
\texttt{EXPORTS\_ATTRIBUTE} & Resource → Attribute & Associates resources with computed output values\\
\addlinespace
\hline
\addlinespace
\texttt{HAS\_EXAMPLE} & Resource → Example & Links resources to usage demonstrations\\
\addlinespace
\hline
\addlinespace
\texttt{REFERENCES}* & Argument → Attribute & Captures dependencies where arguments reference\\
& Argument → Argument & attributes or arguments from other resources\\
\bottomrule
\end{tabular}
\end{table}

The hierarchical nature of the relationships enables the representation of deeply nested configurations, where blocks can recursively contain other blocks and arguments. This structure mirrors Terraform's configuration syntax while making implicit relationships explicit for retrieval and reasoning.

\begin{figure}[t]
 \centering
 \includegraphics[width=0.6\textwidth]{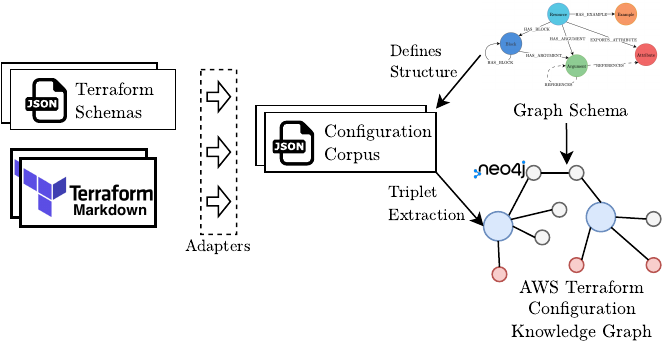} 
 \caption{Knowledge Graph Construction Pipeline Integrating tfschema Structural Definitions with Documentation Content to Produce an Enriched Graph Representation}
 \label{fig:kg_construction}
\end{figure}

\paragraph{Graph Construction Process}
The knowledge graph integrates tfschema's structural definitions with documentation content through a multi-stage pipeline illustrated in Figure~\ref{fig:kg_construction}.

The construction process consists of three stages: schema extraction, documentation integration, and graph instantiation. In the first stage, an adapter processes \texttt{tfschema} JSON files to extract the authoritative structure, including all arguments with types and constraints, nested blocks with cardinality rules, and computed attributes. In the documentation integration stage, a second adapter parses the structured documentation using rule-based extraction to capture descriptions, examples, and semantic content. It implements hierarchical matching logic to accurately associate documentation with schema elements, even in the presence of structural inconsistencies. Table~\ref{tab:extraction-matching-rules} presents the semi-formal rules used in this extraction and matching process. These rules address challenges in Terraform documentation, where configuration arguments can appear at varying nesting depths and exhibit inconsistent formatting or structure. Resource-level rules (R) extract primary resource information; argument (A) and attribute (AT) rules handle the hierarchical nature of configuration options. Example extraction rules (E) preserve practical usage patterns, which are crucial for understanding resource implementation. Matching rules (M) implement logic to align extracted descriptions with schema elements, even when documentation diverges from the schema structure. Schema enrichment rules (S) ensure clean integration of extracted content. The adapter applies all these rules to produce a unified JSON structure conforming to the defined graph schema. To assess the completeness of the enrichment process, we measured coverage for the extracted schema elements. Out of 1875 top-level arguments, 1531 (81.7\%) were matched with descriptive documentation. For block-level arguments, 1875 out of 2519 (74.4\%) were matched with descriptions. For attributes, 503 out of 527 (95.4\%) successfully matched. In total, the system filled 3406 out of 4394 documentation fields (77.5\%). In the final graph Instantiation stage, the enriched schemas are transformed into graph triplets and loaded into Neo4j~\footnote{\url{https://neo4j.com.}}. Neo4j is chosen for its native graph database model and its Cypher query language, which enables efficient traversal during retrieval operations. Each resource is modeled as a central node, connected to entities via the defined relationships. The final graph contains 6519 nodes and 13764 edges, covering all 199 resources in the AWS provider.

\begin{table}[t]
\centering
\footnotesize
\caption{Semi-Formal Rules for Documentation Extraction and Schema Matching}
\begin{tabular}{|l|p{9.5cm}|}
\hline
\textbf{Rule Type} & \textbf{Rule} \\
\hline
[R] Resource Extraction & IF line matches \texttt{"\# Resource: <name>"}, THEN extract \texttt{name} AND find description between title and first \texttt{"\#\# section"}. \\
\hline
[A] Argument Identification & IF in \texttt{Argument Reference} section AND line matches \texttt{"- `<arg\_name>`"}, THEN extract \texttt{arg\_name} AND capture following text until next argument or section. \\
\hline
[A] Section Context & IF line matches \texttt{"\#\#\# <section\_name>"} within \texttt{Argument Reference}, THEN set \texttt{section\_context} for subsequent arguments. \\
\hline
[E] Example Block Detection & IF in \texttt{Example Usage} section AND text contains \texttt{"```"}, THEN extract code block AND preceding title as example. \\
\hline
[E] Example Title Extraction & IF line matches \texttt{"\#\#\# <title>"} before code block, THEN use \texttt{title}; ELSE generate \texttt{"Basic Usage"} or \texttt{"Example N"}. \\
\hline
[AT] Attribute Extraction & IF in \texttt{Attribute Reference} section AND line matches \texttt{"- `<attr\_name>` - <description>"}, THEN extract \texttt{attr\_name} AND \texttt{description}. \\
\hline
[M] Section Normalization & Normalize \texttt{section\_name} by: lowercase → remove \texttt{"block"}, \texttt{":"} → replace spaces with \texttt{"\_"}. \\
\hline
[M] Top-Level Matching & IF \texttt{section\_context} = \texttt{"top-level"} AND schema contains argument \texttt{arg\_name}, THEN assign description directly. \\
\hline
[M] Block Argument Matching & IF \texttt{section\_context} matches normalized block name in schema AND schema block contains \texttt{arg\_name}, THEN assign description to nested argument. \\
\hline
[M] Hierarchical Fallback & IF direct match fails, THEN try: parent block context → combined section paths → top-level → name-based section matching. \\
\hline
[M] Path-Based Matching & FOR nested blocks, construct \texttt{block\_path} = \texttt{[parent, child, ...]} AND match using \texttt{section\_path} with path hierarchy. \\
\hline
[S] Schema Enrichment & IF argument in schema AND matching description found, THEN \texttt{schema.argument.description} = \texttt{cleaned\_description}. \\
\hline
[S] Description Cleaning & Apply \texttt{clean\_description()}: remove excess whitespace → preserve backticks and requirement indicators → trim leading/trailing spaces. \\
\hline
\end{tabular}
\label{tab:extraction-matching-rules}
\end{table}

This process constructs a knowledge graph that combines the structural precision of provider schemas with the semantic richness of documentation, enabling accurate retrieval and context understanding during code generation.

\subsection{Configuration Knowledge Injection Methods}
\label{sec:config_knowledge_methods}
The experimental design will first establish a robust baseline by comparing two fundamental RAG architectures. Following this initial comparison, the more effective RAG architecture will serve as the foundation for a series of iterative enhancements. Each subsequent experiment will introduce a specific type of configuration knowledge or a refined retrieval mechanism, and its impact on code generation will be evaluated.

\subsubsection{Naive RAG Implementation}
\label{ssec:naive_rag}
\begin{figure}[t]
   \centering
   \includegraphics[width=\linewidth]{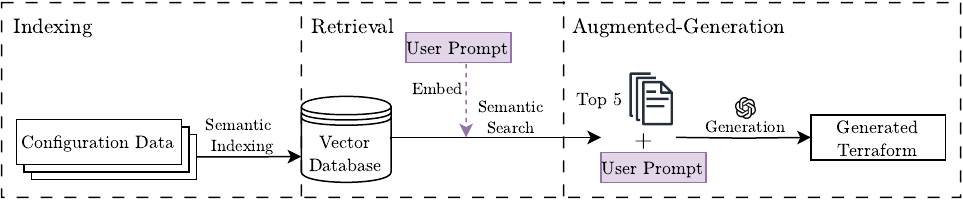}
   \caption{Naive RAG Pipeline Showing the Three-Phase Workflow: indexing of configuration data into vector embeddings, semantic retrieval of top-5 relevant chunks, and augmented generation combining retrieved context with user prompts.}
   \label{fig:naive_flow}
\end{figure}

RAG enhances LLM capabilities by grounding generation in external knowledge sources, reducing reliance on potentially outdated or incomplete internal knowledge\cite{lewis2020retrieval, borgeaud2022improving}. The most fundamental implementation of this paradigm is Naive RAG, which follows a three-phase workflow: indexing, retrieval, and augmented generation\cite{ma2023query}, as illustrated in Figure~\ref{fig:naive_flow}. 

During the \textbf{indexing phase}, the semantic embeddings creation process established in Section~\ref{ssec:sem-emb} is leveraged. Terraform documentation chunks are pre-processed and embedded using a 768-dimensional vector generated by the \texttt{sentence-transformers/all-mpnet-base-v2} model.  

In the \textbf{retrieval phase}, query processing begins with embedding the user request using the same sentence-transformer model used during indexing. The embedded query vector is matched against all stored document embeddings using cosine similarity. The system retrieves the top 5 most similar chunks, ranked in descending order of similarity score. This strategy prioritizes simplicity and reproducibility in line with established RAG implementation guidelines~\cite{lewis2020retrieval}. No similarity thresholds, advanced reranking mechanisms, or query expansion techniques are applied, ensuring that the baseline reflects standard practices without optimization artifacts that could confound comparative analysis~\cite{gao2023retrieval}. 

The \textbf{augmented generation phase} involves adding retrieved documentation chunks to the LLM prompt using a standardized template. This maintains a clear separation between retrieved context and user instructions:
\begin{lstlisting}[basicstyle=\footnotesize\ttfamily, frame=single]
Here is additional knowledge retrieved from Terraform documentation that may help answer the question:
DOCUMENTATION: {context}
USER QUERY: {query_text}
Generate the appropriate Terraform code to address the query.
\end{lstlisting}
The retrieved chunks are concatenated in order of similarity rank, with the most relevant documentation appearing first in the context window, following the most straightforward Naive RAG approach~\cite{lewis2020retrieval}. This integration method serves as a baseline for evaluating the impact of structured knowledge representation explored in subsequent graph-based methods. 

\subsubsection{Graph RAG Implementation}
\label{ssec:graph_rag}
Graph Retrieval-Augmented Generation (Graph RAG) evolves the RAG paradigm to address limitations in traditional vector-based retrieval for structured and interconnected knowledge~\cite{procko2024graph, peng2024graph}. While Naive RAG treats documentation as unstructured text chunks, Graph RAG leverages the explicit relationships in the knowledge graph to provide more comprehensive, structurally coherent configuration information.

\begin{figure}[t]
   \centering
   \includegraphics[width=\linewidth]{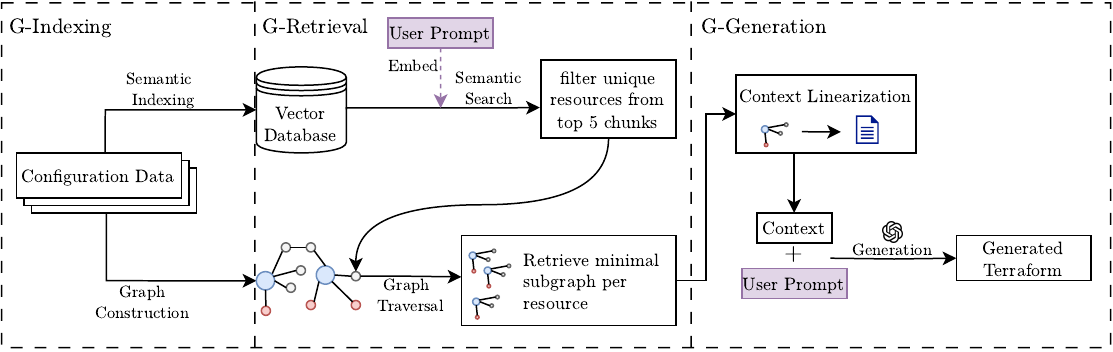}
   \caption{Graph RAG Pipeline Demonstrating the Integration of Semantic Search with Graph Traversal. The system combines vector-based resource identification with structured graph retrieval to provide a comprehensive configuration context.}
   \label{fig:graph_rag_flow}
\end{figure}

Graph RAG follows the same three-phase architecture as Naive RAG but adapts each phase to work with graph-structured knowledge: G-Indexing, G-Retrieval, and G-Generation~\cite{peng2024graph}. Figure~\ref{fig:graph_rag_flow} illustrates the complete pipeline, highlighting the key differences from the vector-only approach.

The \textbf{Graph-based Indexing (G-Indexing)} maintains the same semantic embedding creation process used in Naive RAG to ensure fair comparison between approaches. The knowledge graph constructed in Section~\ref{ssec:knowledge_graph} serves as the structured knowledge repository, with nodes and relationships pre-loaded into the graph database. This parallel indexing strategy ensures that both retrieval mechanisms operate on identical source documentation while enabling different access patterns.

\textbf{Graph-guided retrieval (G-Retrieval)} transforms semantic search results into structured configuration knowledge through a two-stage process. Initially, it performs an identical semantic search to Naive RAG, embedding the user query and retrieving the top-5 most similar documentation chunks. However, instead of directly using these chunks as context, Graph RAG extracts the unique resource names from the chunk metadata to identify relevant graph entry points.

The second stage performs a structured traversal of the knowledge graph for each identified resource. The traversal strategy implements a constrained subgraph extraction designed to capture required configuration elements:

\begin{lstlisting}[basicstyle=\footnotesize\ttfamily, frame=single]
MATCH (r:Resource {name: $resource_name})
OPTIONAL MATCH (r)-[:HAS_ARGUMENT]->(arg:Argument)
WHERE arg.required = true
OPTIONAL MATCH (r)-[:HAS_ARGUMENT]->(opt_arg:Argument) 
WHERE opt_arg.required = false
OPTIONAL MATCH (r)-[:HAS_BLOCK]->(b:Block)
WHERE b.cardinality[0] >= 1
OPTIONAL MATCH (r)-[:HAS_EXAMPLE]->(e:Example)
WHERE e.index = 0
RETURN r, arg, opt_arg, b, e
\end{lstlisting}

This Cypher query extracts a minimal subgraph centered on each Terraform resource node, including required arguments with their type constraints, optional arguments for completeness, required blocks with their nested structures, and the primary usage example. The traversal implements depth control by limiting expansion to only the required blocks, thereby preventing exponential growth in complex, nested configurations.

In the \textbf{Graph-Enhanced Generation (G-Generation) phase}, the retrieved graph elements are transformed into linear text suitable for LLM processing, preserving structural relationships while organizing information hierarchically. This context linearization process follows a standardized template that prioritizes the required configuration elements:

\begin{lstlisting}[basicstyle=\footnotesize\ttfamily, frame=single]
RESOURCE: {resource_name}
Description: {resource_description}
REQUIRED ARGUMENTS:
- {arg_name} ({type}): {description}
OPTIONAL ARGUMENTS:  
- {arg_name} ({type})
REQUIRED BLOCKS:
{block_name} (cardinality: {min}-{max}):
 - {nested_arg_name} ({type}): {description}
BASIC USAGE EXAMPLE:
{example_code}
\end{lstlisting}

The context formation process uses several techniques to optimize information density. Parent-child relationships are preserved through consistent indentation, reflecting the hierarchical nature of Terraform configurations. Priority ordering places required arguments and blocks before optional elements, ensuring that essential configuration information appears early in the context window. Redundant relationship labels are consolidated to reduce verbosity while maintaining structural clarity.

\subsubsection{Enhancements to the Graph RAG Pipeline}
\label{ssec:graph_rag_enhancements}
Following the baseline comparison, Graph RAG demonstrated superior performance over Naive RAG, establishing it as the foundation for systematic enhancements. We present these results in Section \ref{sec:eval}. The enhancement framework implements targeted improvements to address specific limitations identified in the baseline evaluation. Each enhancement is designed to reduce identified error patterns from our NL2IAC error taxonomy while maintaining compatibility with the core Graph RAG architecture.

\begin{figure}[t]
   \centering
   \includegraphics[width=\linewidth]{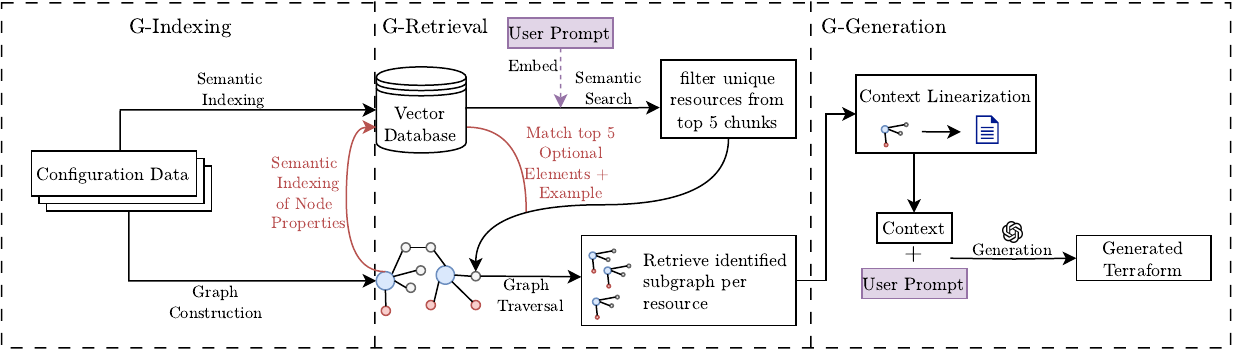}
   \caption{GR-OptMatch Architecture with Semantic Enrichment of Optional Elements. The system performs dual-level semantic matching: document-level for resource identification and node-level for selecting relevant optional arguments and examples.}
   \label{fig:semantic_enrichment}
\end{figure}

\paragraph{Enhancement I: GR-OptMatch - Semantic Enrichment of Optional Elements}
\label{sssec:semantic_enrichment}
The baseline Graph RAG implementation relies on semantic search over documentation chunks to identify relevant resources, followed by deterministic graph traversal that retrieves only required arguments and blocks. While this approach ensures completeness for mandatory configuration elements, it systematically excludes optional configurations that may be highly relevant to specific user requirements, potentially leading to incomplete or suboptimal generated code. To address this behavior, we extended the semantic indexing approach to individual graph nodes, specifically targeting optional arguments, blocks, and usage examples. These are extracted from their respective graph nodes and embedded using the \texttt{sentence-transformers/all-mpnet-base-v2} model during the indexing phase, as shown in Figure~\ref{fig:semantic_enrichment}. The enhanced retrieval process extends the baseline workflow: First, relevant resources are identified through document-level semantic search; second, for each identified resource, the system performs node-level semantic queries to select the most contextually relevant optional elements, ranking them by their semantic similarity to the user query. The top 5 most relevant optional arguments and blocks are selected, along with the most semantically similar usage example.

The Cypher query for graph traversal requires minimal modification to accommodate semantically selected elements: 

\begin{lstlisting}[basicstyle=\footnotesize\ttfamily, frame=single]
MATCH (r:Resource {name: $resource_name})
OPTIONAL MATCH (r)-[:HAS_ARGUMENT]->(arg:Argument)
WHERE arg.required = true
OPTIONAL MATCH (r)-[:HAS_ARGUMENT]->(opt_arg:Argument) 
WHERE opt_arg.required = false AND opt_arg.name IN $selected_optional_args
OPTIONAL MATCH (r)-[:HAS_BLOCK]->(b:Block)
WHERE b.cardinality[0] >= 1 OR b.name IN $selected_optional_blocks
OPTIONAL MATCH (r)-[:HAS_EXAMPLE]->(e:Example)
WHERE e.name = $selected_example_title
RETURN r, arg, opt_arg, b, e
\end{lstlisting}
The query modification introduces parameter-driven filtering for optional elements, where \texttt{\$selected\_optional\_args}, \texttt{\$selected\_optional\_blocks}, and \texttt{\$selected\_example\_title} contain the results of semantic matching performed against the embedded node descriptions.

\paragraph{Enhancement II: GR-LLMSum - LLM-Enhanced Descriptions}
\label{sssec:gr_llmsum}
The baseline Graph RAG implementation and GR-OptMatch enhancement both rely on existing textual descriptions from Terraform documentation for semantic matching. The available descriptions often use technical terminology that may not align with natural language queries from users seeking specific functionality. In addition, technical documentation frequently features sparse, terse descriptions that prioritize brevity over semantic richness. In the Terraform documentation, this sparsity manifests in several ways:\textbf{(i) Terminology Gaps:} Official documentation uses precise technical terms that may not match user intent. \textbf{(ii) Missing Context:} During the construction of the KG, only 77.5\% of descriptions could be matched with their corresponding elements. A quarter of the elements in the KG lack semantic context beyond their element names. \textbf{(iii) Limited Semantic Density:} Existing descriptions often focus on structural information rather than functional capabilities, reducing their effectiveness for semantic similarity matching.

GR-LLMSum addresses these limitations by generating semantically rich descriptions for all knowledge graph entities using targeted LLM prompts~\cite{YANG2024112155}. The enhancement operates during the knowledge graph construction phase, supplementing or replacing sparse descriptions with LLM-generated summaries optimized for semantic search.
The system utilizes distinct prompt templates for each entity type, designed to extract the maximum semantic value while maintaining technical accuracy. Each prompt instructs the LLM to focus on practical capabilities, common use cases, and the functional outcomes users can achieve. For resource entities, the prompt emphasizes primary capabilities and common configuration scenarios, and for argument entities, the prompt connects configuration parameters to user goals and functional outcomes. GR-LLMSum utilizes the same architecture as the GR-OptMatch implementation (Figure \ref{fig:semantic_enrichment}). Instead of the textual description of the elements, their LLM-generated summaries are embedded.

\begin{figure}[t]
   \centering
   \includegraphics[width=\linewidth]{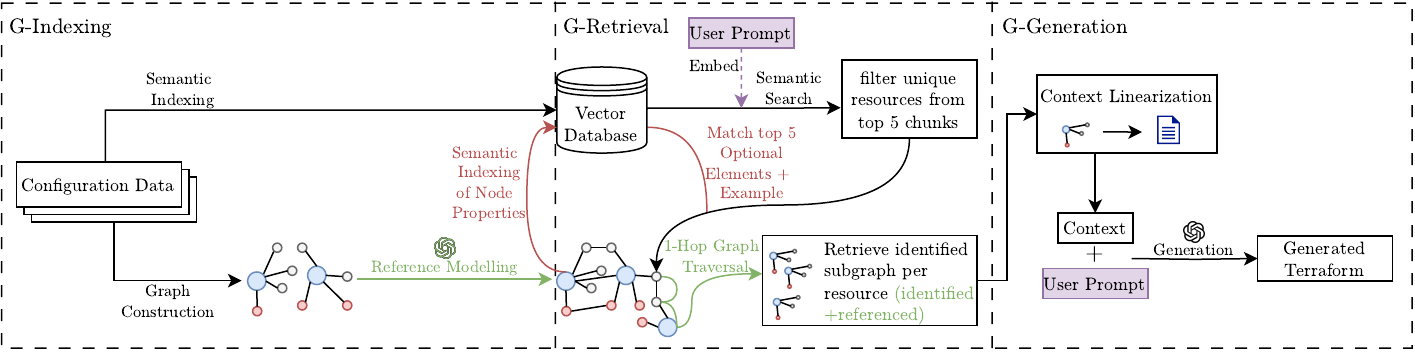}
   \caption{GR-Ref Architecture Showing the Addition of Reference Relationship Modeling and 1-hop Graph Traversal (Green). The system extracts cross-resource dependencies using LLMs and performs extended traversals to automatically include referenced resources in the retrieval context.}
   \label{fig:gr_ref_architecture}
\end{figure}

\paragraph{Enhancement III: GR-Ref - Inter-Resource Dependency Modeling}
\label{sssec:gr_ref}
Analysis of generation failures revealed a persistent issue: errors occurring when generated configurations required dependencies between multiple resources that were not identified during the initial semantic search phase.  These missing dependencies often involve implicit relationships where one resource's arguments must reference outputs from other resources, leading to incomplete or invalid multi-resource configurations. Terraform configurations frequently require explicit references between resources through argument-attribute relationships. When the initial semantic search identifies only a subset of the required resources, the generated configuration lacks these critical dependencies, resulting in syntactically correct but functionally incomplete infrastructure code. GR-Ref leverages this capability by using LLMs to extract inter-resource dependency relationships directly from Terraform documentation, which is often implicit. The approach analyzes each resource's arguments to identify arguments that must reference outputs from other resources, creating explicit REFERENCES edges in the knowledge graph. The extraction process uses a prompt designed to identify cross-resource dependencies. Figure~\ref{fig:gr_ref_architecture} illustrates the enhanced architecture, which incorporates cross-resource relationship modeling and 1-hop graph traversal. This systematic extraction process serves as an additional step after the initial knowledge graph construction, analyzing the entire documentation corpus to identify implicit dependency relationships.

\begin{figure}[t]
   \centering
   \includegraphics[width=\linewidth]{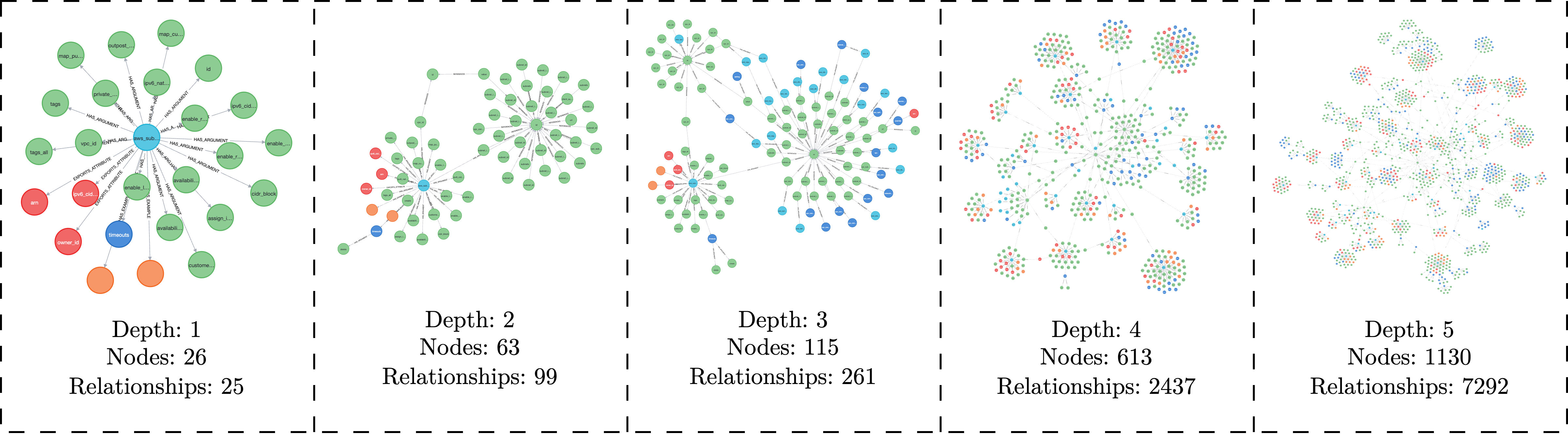}
   \caption{Graph Traversal Starting from \texttt{aws\_subnet} Resource.}
   \label{fig:traversal_depth}
\end{figure}

As shown in Figure~\ref{fig:traversal_depth}, the extraction process achieved substantial coverage across the Terraform AWS provider documentation, expanding exponentially with depth. This growth would quickly overwhelm LLM context windows, introducing irrelevant information and highlighting the importance of limiting traversal depth in graph-based retrieval. The 1-hop approach (depth 2) provides an optimal balance, capturing immediate cross-resource dependencies while maintaining computational feasibility and context relevance.

\subsection{Experimental Design and Methodology}
\label{sec:method_experimental_design}
This research employs a comparative experimental design with incremental enhancement evaluation to systematically investigate how different approaches to configuration knowledge injection affect the quality of LLM-generated Terraform code~\cite{wohlin2012experimentation}. The experimental framework addresses \textit{RQ2} through controlled comparisons that isolate specific variables while maintaining consistency across all other experimental conditions, following established practices in RAG research~\cite{borgeaud2022improving, siriwardhana2023improving} and in graph-based knowledge enhancement studies~\cite{dong2024advanced, he2024g}. The experimental strategy employs a progression of knowledge-injection methods, with each subsequent experiment building on the findings of the previous one. This incremental approach enables identifying which specific enhancements contribute most to improved code generation quality, while avoiding confounding effects that might arise from the simultaneous implementation of multiple interventions.

\subsubsection{Experimental Procedure}
\label{ssec:experimental_procedure}

A consistent experimental procedure was applied across all interventions designed to inject configuration knowledge and improve LLM-generated Terraform code. This procedure ensures that each distinct method is evaluated under comparable conditions using the standardized validation benchmark detailed in Section~\ref{sec:benchmark}.

For each experimental condition or configuration knowledge injection method tested, the following steps were performed for every prompt within the benchmark:

First, we presented the user prompt from the benchmark to the LLM that was configured with the specific knowledge injection method under evaluation. The designated generation LLM then produced the Terraform code, conditioned on both the user prompt and the integrated configuration knowledge. The generated Terraform script was subsequently subjected to the automated technical validation process as specified in the benchmark. Scripts successfully passing technical validation then underwent intent validation to assess semantic correctness. For each generation attempt, relevant data were logged, including pass/fail status for both validation stages, error messages, token counts, and any other pertinent information specific to the intervention being tested.

This standardized procedure ensures that all evaluated methods for injecting configuration knowledge are assessed on an equivalent basis, using the same set of prompts and validation criteria. The error taxonomy established in Section~\ref{sec:error_taxonomy} serves as the analytical lens for qualitative examination of failures, allowing for a deeper understanding of how different interventions impact observed error patterns.

\subsubsection{Control Variables and Experimental Controls}
\label{ssec:control_variables}
The experimental framework ensures control of variables through several mechanisms. All experiments use identical source documentation derived from the official Terraform AWS provider documentation, ensuring that the knowledge content remains constant while only the representation and retrieval methods vary. The generation model GPT-4o was selected as the flagship model of OpenAI, with the highest score on the initial IaC evaluation benchmark, and was used consistently across all experiments, as this research focuses on configuration knowledge methods rather than model comparison. The temperature parameter was set to 0.0 to ensure deterministic and reproducible generation across all experimental conditions.

Context length consistency was maintained for the baseline comparison between Naive RAG and Graph RAG. Context token lengths were statistically equivalent, confirmed through t-test analysis (t-statistic: -1.57, p-value: 0.117). Since the p-value exceeds the significance threshold of 0.05, there is no statistically significant difference in context length between these baseline approaches, ensuring fair comparison.

Each experiment processes the identical set of benchmark prompts, enabling direct statistical comparison of outcomes across different knowledge injection approaches. Evaluation consistency is maintained through standardized validation procedures applied uniformly across all experimental conditions, with technical validation assessing syntactic correctness and schema compliance, and intent validation evaluating semantic correctness through systematic comparison against expected infrastructure requirements.

\subsubsection{Evaluation Metrics}
\label{ssec:evaluation_metrics}
Performance assessment across all experiments employs three primary metrics designed to capture different aspects of generation quality. Importantly, these metrics are grounded in functional and semantic correctness rather than superficial textual similarity. 

Conventional metrics such as BLEU and CodeBERTScore, which measure surface-level similarity via n-gram or embedding comparisons, are inadequate for evaluating IaC. Declarative languages like Terraform allow flexible argument and resource ordering, meaning functionally correct scripts may differ syntactically. As shown in the IaC-Eval benchmark~\cite{kon2024iac}, these metrics often assign similar scores to both correct and incorrect outputs. This motivates the use of technical- and intent-based validation tailored to the functional correctness of IaC configurations.

\begin{description}
    \item[Technical Validation Pass Rate]  
This metric measures the percentage of generated scripts that pass both \texttt{terraform init} and \texttt{terraform plan}. It ensures syntactic validity and schema-level compliance with the IaC provider.

\item[Intent Validation Pass Rate]  
Among scripts that pass technical validation, this metric evaluates whether the generated configuration aligns with the user’s intended infrastructure goals, utilizing a policy-as-code specification (via OPA). This captures semantic and functional correctness.

\item[Average Context Tokens]  
Average Context Tokens quantifies the mean number of tokens in the contextual information provided to the generation LLM, ensuring comparability across the baseline comparison and enabling analysis of efficiency trade-offs between performance improvements and computational overhead.

\end{description}

Results are displayed in standardized performance tables (Table~\ref{tab:results_template}) with statistical significance analysis (Section~\ref{ssec:statistical_analysis}).

\begin{table}[t]
    \centering
    \caption{\textbf{Results Template:} Performance Comparison Format}
    \label{tab:results_template}
    \footnotesize
    \setlength{\tabcolsep}{12pt}
    \begin{tabular}{lcc}
        \toprule
        \textbf{Metric} & {\textbf{Method A}} & {\textbf{Method B}} \\
        \midrule
        \rowcolor{gray!10}
        TV Pass Rate (\%) & {$\text{TV}_A$} & {$\text{TV}_B$} \\
        \rowcolor{gray!10}
        \quad (N passed TV / N initial prompts) & {($N^{\text{tv}}_A$/$N_{\text{total}}$)} & {($N^{\text{tv}}_B$/$N_{\text{total}}$)} \\
        \addlinespace[0.5em]
        
        IV Pass Rate (on TV passes) (\%) & {$\text{IV}_A$} & {$\text{IV}_B$} \\
        \quad (N passed IV / N passed TV) & {($N^{\text{iv}}_A$/$N^{\text{tv}}_A$)} & {($N^{\text{iv}}_B$/$N^{\text{tv}}_B$)} \\
        \addlinespace[0.5em]
        
        \rowcolor{gray!10}
        Overall Success Rate (\%) & {$\text{Overall}_A$} & {$\text{Overall}_B$} \\
        \rowcolor{gray!10}
        \quad (N passed TV\&IV / N initial prompts) & {($N^{\text{overall}}_A$/$N_{\text{total}}$)} & {($N^{\text{overall}}_B$/$N_{\text{total}}$)} \\
        \addlinespace[0.5em]
        
        Average Context Tokens & {$\text{Tokens}_A$} & {$\text{Tokens}_B$} \\
        \quad (Standard Deviation) & {($\pm\sigma_A$)} & {($\pm\sigma_B$)} \\
        \bottomrule
    \end{tabular}
    \vspace{0.5em}
    \begin{minipage}{\linewidth}
    \centering
      \footnotesize{\textit{Note:} $N_{\text{total}} = 457$ benchmark prompts.}
    \end{minipage}
     
\end{table}

\subsubsection{Statistical Analysis Methods}
\label{ssec:statistical_analysis}

Statistical hypothesis tests are conducted at a significance level of $\alpha = 0.05$, with p-values at or below this threshold indicating statistically significant results. For performance metrics that represent binary outcomes (pass/fail) evaluated on the same set of benchmark prompts under different experimental conditions, McNemar's test~\cite{mcnemar1947note} is employed. This test is appropriate for paired nominal data, with the test statistic calculated as:

\begin{equation}
\chi^2 = \frac{(b-c)^2}{b+c}
\end{equation}

where $b$ and $c$ represent counts of discordant pairs following a chi-squared distribution with one degree of freedom.

When comparing multiple conditions on paired binary data, pairwise McNemar's tests are conducted between all method combinations. For multiple pairwise comparisons, a Bonferroni correction is applied to adjust the significance level and control the family-wise error rate~\cite{bonferroni1936teoria}.

For Intent Validation comparisons, applicable only to prompts passing technical validation, a matched-subset approach addresses potential selection bias. For each pairwise comparison, only prompts that passed Technical Validation under both conditions are retained, with McNemar's test applied to this paired subset and odds ratios reported to quantify effect size. This approach maintains the paired nature of the experimental design while eliminating selection bias.

Error type analysis based on the established taxonomy is primarily descriptive, reporting frequencies and percentages to provide qualitative insights into how different knowledge injection methods affect specific failure patterns. Context token analysis is similarly reported descriptively, using means and standard deviations, to assess computational efficiency across different approaches.

For a comprehensive visual comparison of all methods simultaneously, Critical Difference (CD) diagrams will be employed to illustrate the performance ranking and statistical relationships between methods~\cite{demvsar2006statistical}. While traditional CD diagrams are based on Friedman test rankings across multiple datasets, our adaptation uses performance-based rankings from a single benchmark dataset combined with pairwise McNemar test results to determine statistical equivalence between methods. Methods are visually connected in the diagram when their pairwise McNemar test indicates no statistically significant difference, providing an intuitive overview of the performance landscape across all knowledge injection approaches. To perform the analysis and draw the CD diagram, we adapted the implementation provided by  Ismail Fawaz et al.~\cite{ismail2019deep}.

\section{Evaluation of Configuration Knowledge Injection }\label{sec:eval}

To answer RQ2, we evaluated the techniques we developed for injecting configuration knowledge into the Terraform code generation process. The evaluation involves three phases: 1) establishing performance baselines using No RAG (LLM-only), Naive RAG, and Foundational Graph RAG (\ref{sec:baseline_comparison}); 2) exploring targeted enhancements with Graph RAG strategies—retrieval scope expansion (GR-OptMatch), semantic enrichment (GR-LLMSum), and dependency modeling (GR-Ref) (\ref{sec:graph_enhancement_strategies}); and synthesizing findings to identify optimal strategies and challenges (\ref{sec:overall_synthesis}).

\subsection{Baseline Comparison: Knowledge Structure Impact}
\label{sec:baseline_comparison}
The baseline comparison of No RAG (LLM-only), Naive RAG, and Foundational Graph RAG revealed clear performance distinctions. Foundational Graph RAG generally achieved the highest scores across key quality metrics, followed by Naive RAG, with both RAG techniques surpassing the No RAG baseline. Table~\ref{tab:baseline_performance_summary} summarizes the results.

\begin{table}[t]
    \centering
    \caption{\textbf{Performance Summary:} Basline vs. Naive RAG vs. Foundational Graph RAG}
    \label{tab:baseline_performance_summary}
    \footnotesize
    \setlength{\tabcolsep}{10pt} 
    \begin{tabular}{lccc}
        \toprule
        \textbf{Metric} & {\textbf{Base}} & {\textbf{Naive RAG}} & {\textbf{Graph RAG}} \\
        \midrule
        \rowcolor{gray!10} 
        TV Pass Rate (\%) & {37.2} & {70.2} & {80.3} \\
        \rowcolor{gray!10}
        \quad (N passed TV / N initial prompts) & {(170/457)} & {(321/457)} & {(367/457)} \\
        \addlinespace[0.5em] 
        
        IV Pass Rate (on TV passes) (\%) & {72.9} & {74.8} & {72.8} \\
        \quad (N passed IV / N passed TV) & {(124/170)} & {(240/321)} & {(267/367)} \\
        \addlinespace[0.5em] 
        
        \rowcolor{gray!10} 
        Overall Success Rate (TV \& IV) (\%) & {27.1} & {52.5} & {58.4} \\
        \rowcolor{gray!10}
        \quad (N passed TV\&IV / N initial prompts) & {(124/457)} & {(240/457)} & {(267/457)} \\
        \addlinespace[0.5em] 
        
        Average Prompt Tokens & {$175$} & {$1432$} & {$1462$} \\
        \quad (Standard Deviation) & {($\pm 36$)} & {($\pm 300$)} & {($\pm 577$)} \\
        \bottomrule
    \end{tabular}
    \vspace{0.5em} 
    
\end{table}

\subsubsection{Statistical Analysis}
\label{ssec:baseline_statistical_analysis}

\begin{table}[t]
    \centering
    \caption{\textbf{McNemar's Test Results for Pairwise Comparisons}}
    \label{tab:mcnemar_results}
    \footnotesize
    \begin{tabular}{llccc}
        \toprule
        \textbf{Comparison} & \textbf{Test} & $\boldsymbol{\chi^2}$ & \textbf{p-value} & \textbf{Odds Ratio} \\
        \midrule
        \rowcolor{gray!10} 
        Graph RAG vs. Naive RAG & TV & 17.85 & $<$ 0.001 & 2.31 \\
        \rowcolor{gray!10} 
        & Matched IV & 0.01 & 0.923 & 0.93 \\
        \addlinespace[0.3em]
        Graph RAG vs. Base & TV & 162.92 & $<$ 0.001 & 10.85 \\
        & Matched IV & 1.21 & 0.272 & 1.50 \\
        \addlinespace[0.3em]
        \rowcolor{gray!10} 
        Naive RAG vs. Base & TV & 101.57 & $<$ 0.001 & 5.19 \\
        \rowcolor{gray!10} 
        & Matched IV & 6.37 & 0.012 & 2.67 \\
        \bottomrule
    \end{tabular}
    \vspace{0.5em} 
    \begin{minipage}{\linewidth}
    \centering
      \footnotesize{\textit{Note:} Matched IV analysis conducted on prompts passing TV in both conditions.}
    \end{minipage}
\end{table}

\begin{description}
    \item[TV Performance] As shown in Table~\ref{tab:baseline_performance_summary}, the TV pass rates differed substantially across the three approaches: 37.2\% for No RAG, 70.2\% for Naive RAG, and 80.3\% for Graph RAG. A series of pairwise McNemar's tests revealed statistically significant differences in TV pass rates among all three approaches (Table~\ref{tab:mcnemar_results}). Graph RAG significantly outperformed Naive RAG, with an odds ratio of 2.31, indicating that Graph RAG had 2.31 times the odds of producing technically valid code when Naive RAG failed, compared to the reverse scenario. Similarly, Graph RAG significantly outperformed No RAG  with a substantial odds ratio of 10.85. Naive RAG also significantly outperformed No RAG with an odds ratio of 5.19. 
    The pattern of improvements observed in the TV contingency tables indicates that each augmentation approach builds upon the capabilities of the previous one. Of the 457 prompts evaluated, 286 passed TV in both Graph RAG and Naive RAG implementations. Graph RAG successfully handled an additional 81 prompts that Naive RAG failed on, while Naive RAG uniquely passed only 35 prompts that Graph RAG could not process correctly. This demonstrates that Graph RAG largely retains the technical capabilities of Naive RAG while extending them to cover additional use cases.
    \item[IV Performance] The Intent Validation (IV) pass rates on prompts that passed TV were remarkably similar across implementations: 72.9\% for No RAG, 74.8\% for Naive RAG, and 72.8\% for Graph RAG. To assess whether these differences were statistically significant, we conducted McNemar's tests on matched subsets—prompts that passed TV in both conditions being compared. 
    For the Graph RAG vs. Naive RAG comparison on the 286 prompts that passed TV in both implementations, no significant difference in IV performance was found. The odds ratio of 0.93 suggests a slight, though not statistically significant, advantage for Naive RAG in intent alignment. Similarly, for Graph RAG vs. No RAG, no significant difference in IV performance was found across their 150 shared TV passes. Interestingly, Naive RAG did show a statistically significant improvement over No RAG in IV performance on their 134 shared TV passes, with an odds ratio of 2.67.
    These results suggest that while graph-based knowledge representation significantly improves technical correctness, it does not provide a significant advantage in terms of intent alignment once the code is technically valid. This may be partly explained by the fact that the base Graph RAG implementation uses a relatively basic retrieval mechanism that primarily captures required arguments and blocks, and lists only top-level optional arguments without including optional nested blocks. In contrast, Naive RAG's more semantic retrieval might inadvertently capture more intent-related information through its chunk-based approach, potentially offsetting Graph RAG's structural advantages for intent alignment. This observation suggests that there is room for improvement in how Graph RAG handles intent-related information, which will be explored in subsequent experiments.
    \item[Overall Success Rates] The Overall Success Rate, representing prompts that passed both TV and IV, showed a clear progression across the three implementations: 27.1\% for No RAG, 52.5\% for Naive RAG, and 58.4\% for Graph RAG. This substantial improvement in overall performance—more than doubling the success rate from No RAG to Graph RAG—demonstrates the significant practical impact of both Retrieval-Augmented Generation approaches.
    The TV$\rightarrow$IV transition rates (percentage of TV passes that also passed IV) were remarkably consistent across implementations: 72.9\% for No RAG, 74.8\% for Naive RAG, and 72.8\% for Graph RAG. This consistency further supports the finding that the primary difference between approaches lies in their ability to produce technically valid code rather than in aligning that code with user intent.
\end{description}

\subsubsection{Qualitative Error Analysis}
We conducted a detailed qualitative analysis of how various RAG strategies impact LLM generation failure patterns, with a primary focus on Dimension 2 of our error taxonomy (Section \ref{sec:error_taxonomy}): Factual Incorrectness and Incompleteness. This analysis provides deeper insights into the statistical differences observed in our quantitative evaluation.

\begin{table}[t]
     \caption{Total Factual Incorrectness and Incompleteness Errors in Technical Validation across Baseline, Naive RAG, and Graph RAG Approaches. Counts derived from applying the NL2IaC Error Taxonomy to technical errors.}
     \label{tab:comparison_3_dim2}
    \footnotesize
    \begin{tabular}{c|c|c|c}
     & Base & Naive RAG & Graph RAG \\ \hline
    Factual Incorrectness & 502 & 136 & 61 \\ \hline
    Incompleteness & 204 & 44 & 39
    \end{tabular}
\end{table}

\paragraph{TV Error Distribution Patterns}
The baseline model \textbf{No RAG}, which relies solely on the inherent knowledge of GPT-4o, exhibits substantial challenges in generating technically valid IaC. As elaborated upon during the development of our error taxonomy (refer to Section~\ref{sec:error_taxonomy}), technical validation failures within this framework are predominantly attributed to errors of Factual Incorrectness and Incompleteness. Specifically, 326 instances of factual incorrectness and 140 instances of incompleteness were recorded.

Table~\ref{tab:comparison_3_dim2} visually demonstrates the pronounced reduction in these error types following the integration of RAG methodologies. With the introduction of Naive RAG, a substantial decrease is observed: Factual Incorrectness errors decreased by 75\% (from 326 to 81), and Incompleteness errors decreased by 81\% (from 140 to 27). This results in an incorrectness-to-omission ratio of approximately 3:1. The implementation of chunk-based retrieval evidently helps the LLM adhere to valid Terraform structures; however, the persistence of 81 factual incorrectness errors underscores the limitations inherent in how semantic chunks encapsulate schema constraints.

Graph RAG mitigates Factual Incorrectness by reducing these errors by an additional 52\% relative to Naive RAG (from 81 to 39). This reduction highlights the efficacy of structured knowledge representation in preventing the incorporation of invalid elements. Nevertheless, it slightly increased Incompleteness errors (from 27 to 32), shifting the error ratio to approximately 1.6:1. This shift suggests that, as Graph RAG significantly curtails Factual Incorrectness through structured knowledge, Incompleteness becomes more salient among the residual technical errors.

This redistribution of error types can potentially be elucidated by examining the fundamental retrieval mechanism of Base Graph RAG. A critical limitation is observed in its basic retrieval capability: while it accurately identifies the need for an optional \texttt{secondary\_artifacts} block, it omits the requisite \texttt{artifact\_identifier} within it. This implementation primarily focuses on retrieving comprehensive information about top-level required arguments and blocks. However, it only provides names (without detailing internal requirements) for optional elements, leading to potential omission errors when required arguments are nested within these optional blocks.

\paragraph{IV Error Patterns}
Our statistical analysis revealed a noteworthy pattern: while Naive RAG demonstrated a statistically significant enhancement in IV performance relative to the baseline, Graph RAG did not exhibit the same advantage, despite its technical superiority. This observed disconnect between technical correctness and intent alignment suggests that distinct knowledge representation strategies might be optimal for these separate facets of code generation.

An analysis of discordant cases indicates that the graph-based approach is proficient in representing structural information, such as required arguments and blocks. However, it may not effectively capture the semantic nuances necessary for precise intent alignment when compared to Naive RAG's semantic similarity matching approach.

Naive RAG's broader semantic retrieval mechanism occasionally includes contextual examples that more closely align with user intent, even though this method is less precise for technical validation purposes. This phenomenon elucidates why Graph RAG shows marked improvements in technical correctness without corresponding gains in intent alignment. These findings underscore the potential benefits of hybrid approaches that combine structural precision with semantic richness.

\subsection{Graph RAG Enhancement Strategies}
\label{sec:graph_enhancement_strategies}
Building upon the baseline Graph RAG, hereafter referred to as GR-Base, this experiment investigated whether enhancing the retrieval phase to include semantic matching against optional arguments, blocks, and examples would further improve performance. We first compare GR-OptMatch directly with the baseline. After that, we evaluate GR-OptMatch, GR-LLMSum, and finally GR-LLMSum and GR-Ref.

\subsubsection{Statistical Analysis}
\begin{table}[t]
    \centering
    \caption{\textbf{Performance Summary: GR-OptMatch vs. GR-Base vs GR-LLMSum vs GR-Ref}}
    \label{tab:performance_summary_graph}
    \footnotesize
    \setlength{\tabcolsep}{10pt} 
    \begin{tabular}{lcccc}
        \toprule
        \textbf{Metric} & {\textbf{GR-Base}} & {\textbf{GR-OptMatch}} & {\textbf{GR-LLMSum}} & {\textbf{GR-Ref}}\\
        \midrule
        \rowcolor{gray!10} 
        TV Pass Rate (\%) & {80.3} & {84.2} & {83.2} & {80.3}\\
        \rowcolor{gray!10}
        \quad (N passed TV / N initial prompts) & {(367/457)} & {(385/457)} & {(380/457)} & {(367/457)}\\
        \addlinespace[0.5em] 
        
        IV Pass Rate (on TV passes) (\%) & {72.8} & {74.0} & {75.3} & {75.2}\\
        \quad (N passed IV / N passed TV) & {(267/367)} & {(285/385)} & {(286/380)} & {(276/367)}\\
        \addlinespace[0.5em] 
        
        \rowcolor{gray!10} 
        Overall Success Rate (TV \& IV) (\%) & {58.4} & {62.4} & {62.6} & {60.4}\\
        \rowcolor{gray!10}
        \quad (N passed TV\&IV / N initial prompts) & {(267/457)} & {(285/457)} & {(286/457)} & {(276/457)}\\
        \addlinespace[0.5em] 
        
        Average Context Tokens per Prompt & {1462} & {2319} & {2418} & {3517}\\
        \quad (Standard Deviation) & {($\pm$577)} & {($\pm$956)} & {($\pm$949)} & {($\pm$1654)}\\
        \bottomrule
    \end{tabular}
    \vspace{0.5em} 
    
\end{table}

\begin{table}[t]
    \centering
    \footnotesize
    \caption{\textbf{McNemar's Test Results for Pairwise Comparisons}}
    \label{tab:mcnemar_results_gr}
    \begin{tabular}{llccc}
        \toprule
        \textbf{Comparison} & \textbf{Test} & $\boldsymbol{\chi^2}$ & \textbf{p-value} & \textbf{Odds Ratio} \\
        \midrule
        \rowcolor{gray!10} 
        GR-OptMatch vs. GR-Base & TV & 4.50 & 0.034 & 1.72 \\
        \rowcolor{gray!10} 
        & Matched IV & 1.08 & 0.299 & 1.55 \\
        \midrule
        \rowcolor{gray!10} 
        GR-OptMatch vs. GR-LLMSum & TV & 0.52 & 0.471 & 1.29 \\
        \rowcolor{gray!10} 
        & Matched IV & 0.56 & 0.456 & 0.69 \\
        \midrule
        \rowcolor{gray!10} 
        GR-LLMSum vs. GR-Ref & TV & 2.20 & 0.138 & 1.45 \\
        \rowcolor{gray!10} 
        & Matched IV & 0.09 & 0.759 & 0.85 \\
        \bottomrule
    \end{tabular}
    \vspace{0.5em} 
    \begin{minipage}{\linewidth}
    \centering
      \footnotesize{\textit{Note:} Matched IV analysis conducted on prompts passing TV in both conditions.}
    \end{minipage}
\end{table}

\paragraph{TV Performance}

In examining the technical validation (TV) performance of various Graph RAG implementations, significant insights were observed. According to Table~\ref{tab:performance_summary_graph}, GR-OptMatch demonstrated a notable enhancement over GR-Base with an increase in TV pass rate from 80.3\% (\(367/457\)) for GR-Base to 84.2\% (\(385/457\)) for GR-OptMatch. This improvement of 3.9 percentage points was statistically significant, as evidenced by the p-value results of McNemar's test as shown in Table \ref{tab:mcnemar_results_gr}. The odds ratio of 1.72 indicates that GR-OptMatch had a 1.72 times higher probability of producing technically valid code when compared to the instances where the base Graph RAG failed.

Further analysis using contingency tables provides additional clarity on this performance boost. Out of the 457 prompts assessed, 342 were successfully processed by both implementations. GR-OptMatch uniquely addressed an additional 43 prompts that GR-Base could not handle, while GR-Base had a unique success with only 25 prompts that GR-OptMatch failed to process. This net improvement of 18 prompts (3.9\% of the total) suggests meaningful technical benefits from expanding retrieval to include optional elements and examples.

When comparing GR-OptMatch to GR-LLMSum, as depicted in Table~\ref{tab:performance_summary_graph}, it was observed that the TV pass rate for GR-LLMSum stood at 83.2\% (\(380/457\)), whereas GR-OptMatch achieved a slightly higher rate of 84.2\% (\(385/457\)). However, this difference was not statistically significant according to McNemar's test.

In the comparison between GR-LLMSum and GR-Ref, Table~\ref{tab:performance_summary_graph} shows that GR-Ref achieved a TV pass rate of 80.3\% (\(367/457\)), while GR-LLMSum reached an improved rate of 83.2\% (\(380/457\)). Despite a 2.9 percentage-point advantage favoring GR-LLMSum, the difference was not statistically significant under McNemar's test. Contingency table analysis revealed that of the 457 prompts evaluated, 338 passed TV in both implementations. GR-LLMSum uniquely succeeded with an additional 42 prompts compared to GR-Ref, while GR-Ref was unique for 29 prompts that GR-LLMSum could not handle correctly.

In summary, the empirical evidence strongly supports the technical superiority of GR-OptMatch over the baseline model, while indicating minimal differences between GR-LLMSum and both GR-OptMatch and GR-Ref in terms of TV performance.

\paragraph{IV Performance}

The IV analysis pass rates among various Graph RAG implementations revealed nuanced findings. As indicated by the data presented in Table~\ref{tab:performance_summary_graph}, there was a marginal improvement in IV performance from 72.8\% for the base Graph RAG to 74.0\% for GR-OptMatch. To determine whether this difference was statistically significant, McNemar's test was conducted on the matched subset of 342 prompts that passed TV in both implementations. The test results showed no significant statistical difference, although an odds ratio of 1.55 suggests a favorable trend for GR-OptMatch regarding intent alignment.

An analysis of the contingency table for this matched subset revealed that, out of 342 prompts that passed TV in both systems, 244 also passed IV. Unique to GR-OptMatch were 17 additional prompts where base Graph RAG failed IV, while base Graph RAG uniquely succeeded with 11 prompts that GR-OptMatch could not handle correctly. This pattern indicates a potential for modest benefits from incorporating optional arguments and examples into the retrieval process, though statistical significance was not achieved.

When comparing GR-OptMatch to GR-LLMSum, the data shows an IV pass rate of 75.3\% for GR-LLMSum versus 74.0\% for GR-OptMatch. Similarly, the matched-subset analysis showed no statistically significant difference.

In the comparison between GR-LLMSum and GR-Ref, both implementations exhibited nearly identical IV pass rates: 75.2\% for GR-Ref and 75.3\% for GR-LLMSum. The matched-subset analysis involving 338 prompts passing TV in both systems also showed no statistically significant difference. Within this subset, IV performance was closely aligned; GR-LLMSum uniquely succeeded with 11 additional prompts compared to GR-Ref, whereas GR-Ref uniquely passed 13 prompts that GR-LLMSum could not handle.

In summary, while slight improvements in IV pass rates were observed between certain implementations, these differences did not reach statistical significance. The trends suggest potential benefits from expanded retrieval strategies, yet further investigation may be necessary to establish significant advancements in intent alignment capabilities.

\paragraph{Overall Success Rate}

The evaluation of the Overall Success Rate, defined as the proportion of prompts that passed both Technical Validation (TV) and Intent Validation (IV), provides valuable insights into performance improvements across different Graph RAG implementations. When comparing GR-OptMatch to base Graph RAG, there was a noticeable improvement in success rates from 58.4\% (\(267/457\)) for base Graph RAG to 62.4\% (\(285/457\)) for GR-OptMatch. This 4.0 percentage-point increase underscores the practical benefits of integrating optional arguments and examples into the retrieval process, thereby enhancing both technical correctness and intent alignment.

Moreover, the transition rate from TV to IV was slightly higher with GR-OptMatch at 74.0\% than with base Graph RAG at 72.8\%. This suggests that the additional context provided by these enhancements may modestly improve the model's capability to align generated code with user intent once technical correctness is achieved.

In comparing GR-OptMatch with GR-LLMSum, the overall success rates were nearly identical: 62.6\% (\(286/457\)) for GR-LLMSum and 62.4\% (\(285/457\)) for GR-OptMatch.

Finally, the comparison between GR-LLMSum and GR-Ref showed a slight variance in overall success rates: 60.4\% (\(276/457\)) for GR-Ref versus 62.6\% (\(286/457\)) for GR-LLMSum. This 2.2 percentage point difference did not achieve statistical significance, suggesting that while there are observable differences, they may not be substantial enough to warrant significant conclusions.

Overall, the analysis of success rates highlights nuanced improvements in Graph RAG implementations and suggests areas where further optimization could yield more pronounced benefits.

\subsubsection{Qualitative Analysis}

We conducted a quality analysis of the success and failure patterns of GR-OptMatch, GR-LLMSum, and GR-Ref. Due to the similarities between the observations for GR-OptMatch and GR-LLMSum, we present the results for GR-OptMatch and GR-Ref. 

\paragraph{GR-OptMatch}
Comparing GR-OptMatch and GR-Base reveals significant improvements in technical validation, driven by improved semantic matching. We observed that GR-OptMatch retrieves more detailed structural information than GR-Base, addressing critical limitations by including required nested arguments within optional blocks.

\begin{figure}[t]
    \centering
    \begin{subfigure}[b]{0.45\textwidth}
        \centering
        \includegraphics[width=\linewidth]{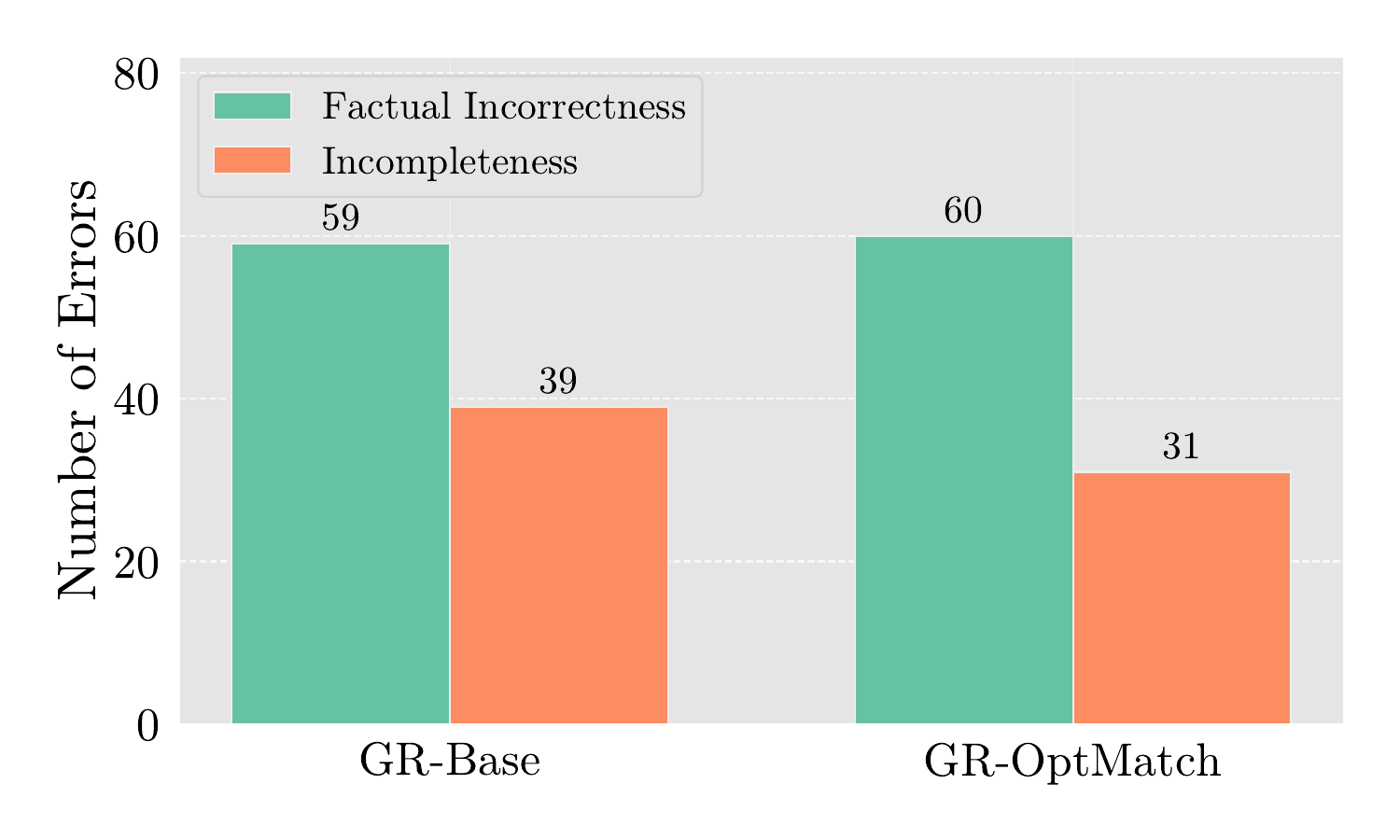}
        \caption{Comparison of Factual Incorrectness (FI) and Incompleteness (IC) Error Counts between GR-Base and GR-OptMatch}
        \label{fig:grb-grom}
    \end{subfigure}
    \hfill
    \begin{subfigure}[b]{0.49\textwidth}
        \centering
        \includegraphics[width=\textwidth]{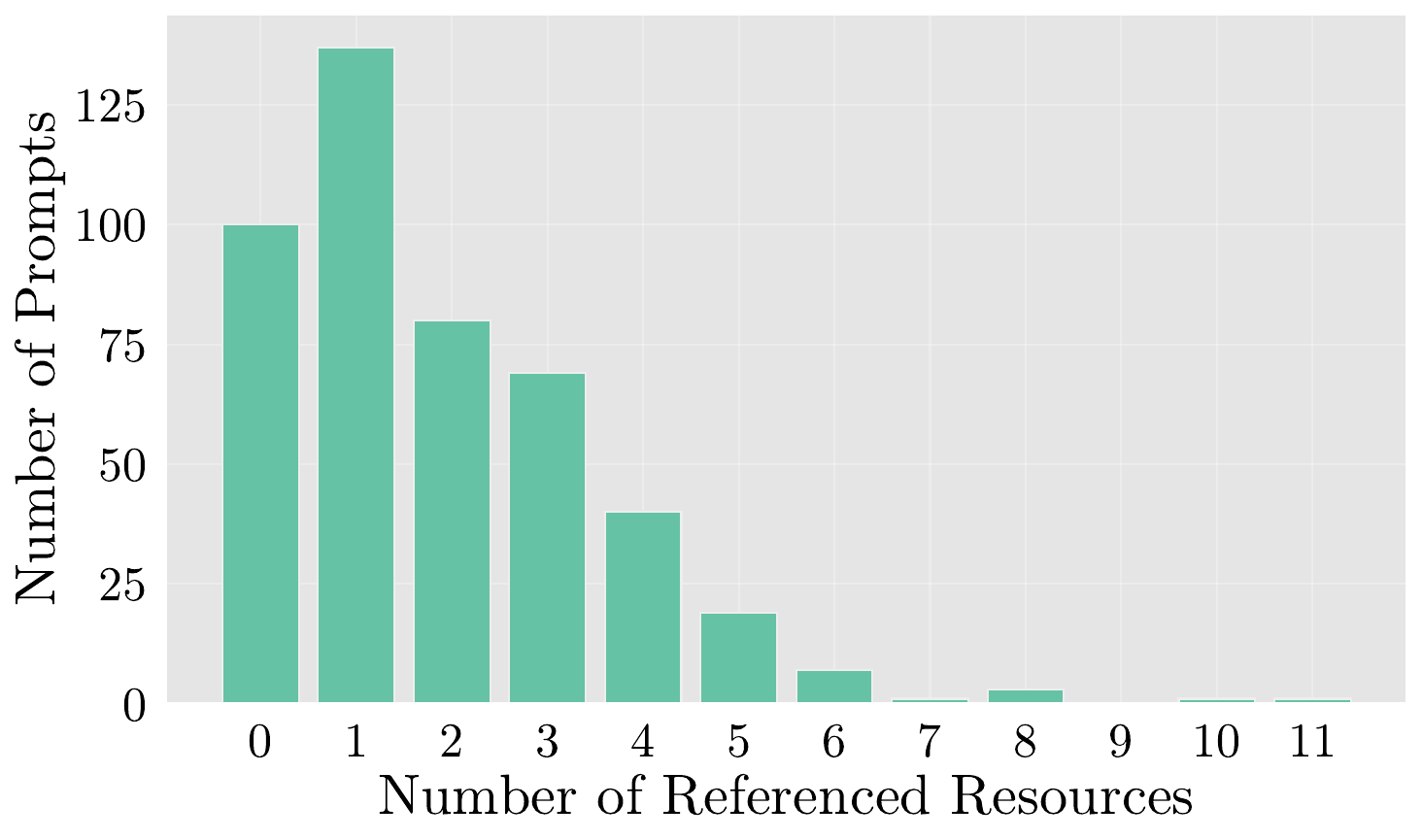}
        \caption{Distribution of Referenced Resources Identified by the GR-Ref Method Across All Prompts}
        \label{fig:referenced_resources_dist}
        \end{subfigure}
        \caption{Baseline Error Statistics}
    \label{fig:baseline_error_stat}
\end{figure}

Figure~\ref{fig:grb-grom} highlights a decrease in Incompleteness errors by 20.5\% (from 39 to 31), while Factual Incorrectness errors slightly increased from 59 to 60. This pattern suggests that semantic matching predominantly reduces omission errors with limited influence on preventing invalid elements.

A review of 10 failed scripts revealed that 70\% of Factual Incorrectness errors originated from unidentifiable resources during retrieval, indicating a persistent bottleneck in resource identification.

Although IV performance improved modestly and was not statistically significant (\(p = 0.299\)), GR-OptMatch maintained comparable intent alignment in complex scripts, uniquely passing technical validation under this model. This contributed to an overall increase from 58.4\% to 62.4\% in scripts passing both TV and IV, suggesting that enhanced retrieval facilitates the generation of more complex configurations without compromising intent alignment.

\paragraph{GR-Ref}

Analyzing the number of referenced resources identified by GR-Ref per prompt reveals important behavioral patterns (Figure~\ref{fig:referenced_resources_dist}). While many prompts involve one or two references, a notable portion involve significantly more, which is both an advantage for complex prompts and a disadvantage for simpler ones due to potential noise.

GR-Ref achieved 9 unique technical validation successes and 12 overall successes (technical and intent validation) where other methods failed. These prompts often involve complex, multi-resource configurations with implicit dependencies not explicitly stated.
A key insight is GR-Ref's ability to detect implicit dependencies, allowing technically valid configurations. This capability extends across various infrastructure setups, successfully managing interdependent resources that other techniques have overlooked.
GR-Ref also captures implicit relationships effectively, fulfilling user intent better than other approaches and identifying essential resource dependencies. For instance, GR-Ref identified essential dependencies for minimal prompts involving \texttt{aws\_connect\_bot\_association} and EKS node group configurations that others missed. 

Despite its strengths, GR-Ref's heuristic of including all detected referenced resources leads to over-complication in simpler tasks. This pattern is evident in unique failures where excessive dependencies introduce unnecessary complexity, resulting in errors.

\subsection{Summary of the Answers to RQ2}
\label{sec:overall_synthesis}
\begin{figure}[t]
    \centering
    \includegraphics[width=0.6\linewidth]{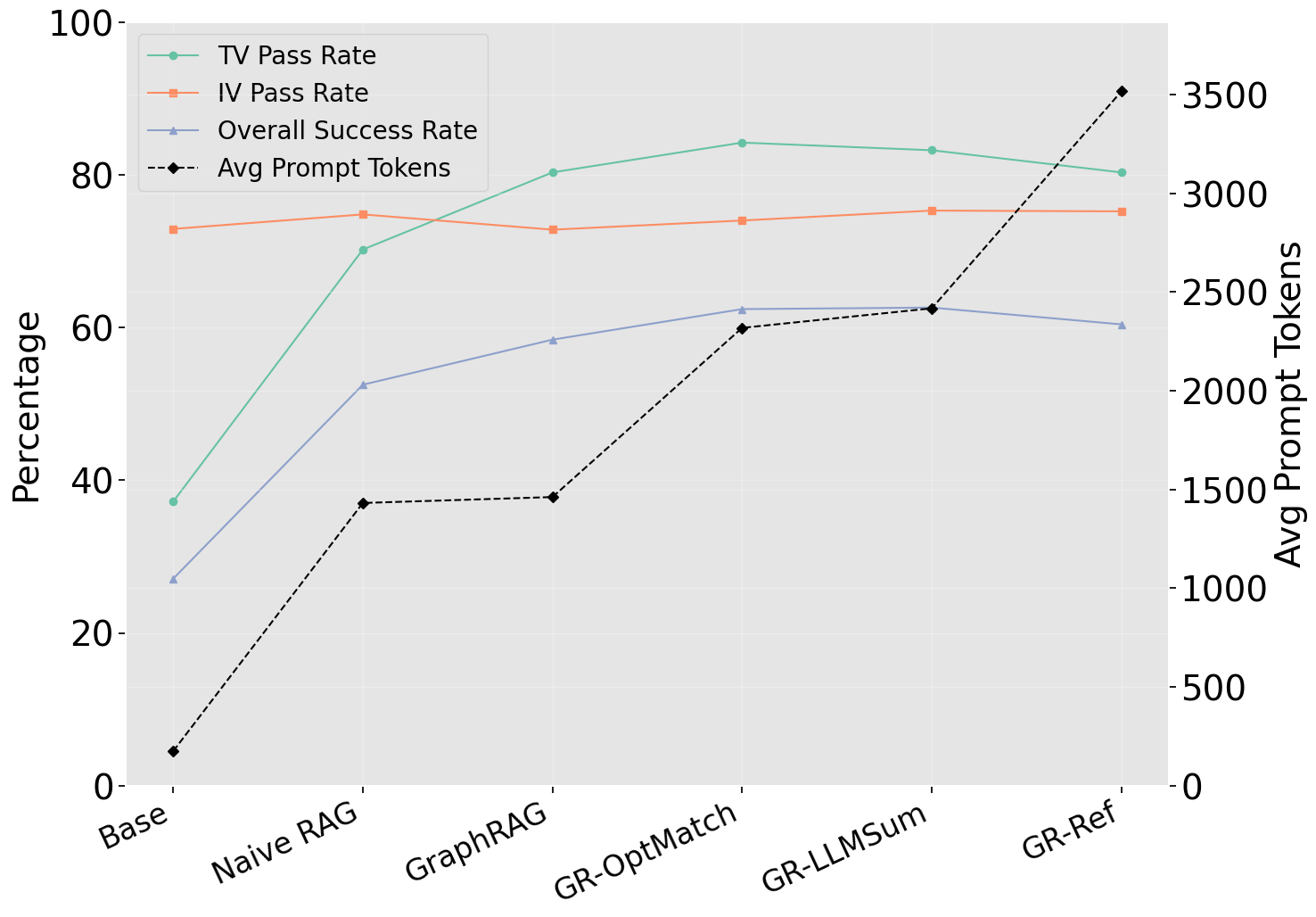}
    \caption{Performance Progression Across All Experimental Methods, including Average Prompt Tokens.}
    \label{fig:progression}
\end{figure}

We developed and evaluated five methods for injecting configuration knowledge into the LLM-based Terraform IaC generation process, namely Naive RAG, Graph-RAG (GR-Base), GR-OptMatch, GR-LLMSum, and GR-Ref. The experimental progression demonstrates a clear enhancement across all knowledge injection approaches, as illustrated in Figure~\ref{fig:progression}. A significant advancement is observed during the initial shift from no knowledge integration (Base) to any form of retrieval-augmented generation, with technical validation rates increasing from 37.2\% to 70.2\% for Naive RAG and from 80.3\% to 90.3\% for Graph RAG. This leap highlights the crucial role of external knowledge injection in IaC generation.

The progression also reveals an important pattern in the relationship between technical validation and intent validation performance. While technical validation rates show substantial variation across methods (ranging from 37.2\% to 93.4\%), intent validation rates remain relatively stable within a narrower band (72.8\% to 76.3\%) once technical validation is achieved. This consistency suggests that intent alignment represents a distinct challenge that is less responsive to the structural and retrieval enhancements that dramatically improve technical correctness.

Analyzing the average prompt tokens provides a perspective on the cost of these performance gains. The initial shift from the baseline to Naive RAG and Graph RAG introduced a significant increase in context size, from 175 tokens to approximately 1,450. The subsequent enhancements, GR-OptMatch and GR-LLMSum, further increased the token count to around 2,400 as more optional context was included. The most substantial increases were seen with GR-Ref (3,517 tokens), driven by dependency inclusion and the iterative repair mechanism's extensive context. This trend underscores a trade-off: while more sophisticated knowledge-injection strategies and adaptive mechanisms deliver higher success rates, they require a significantly larger context window, with direct implications for computational resources.

\begin{figure}[t]
    \centering
    \includegraphics[width=0.8\linewidth]{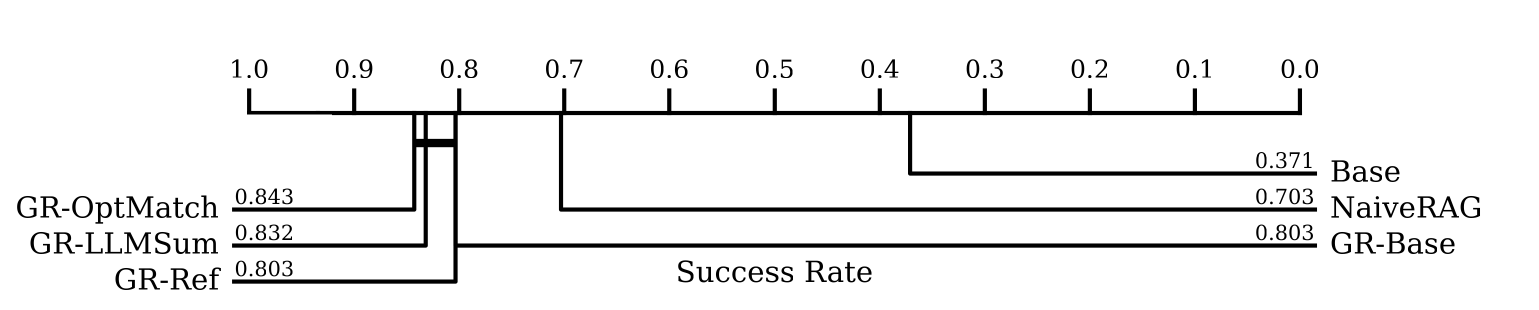}
    \caption{Critical Difference Diagram for Technical Validation Success Rates. Methods connected by horizontal lines show no statistically significant difference based on pairwise McNemar tests.}
    \label{fig:tv_success}
\end{figure}

\begin{figure}[t]
    \centering
    \includegraphics[width=0.8\linewidth]{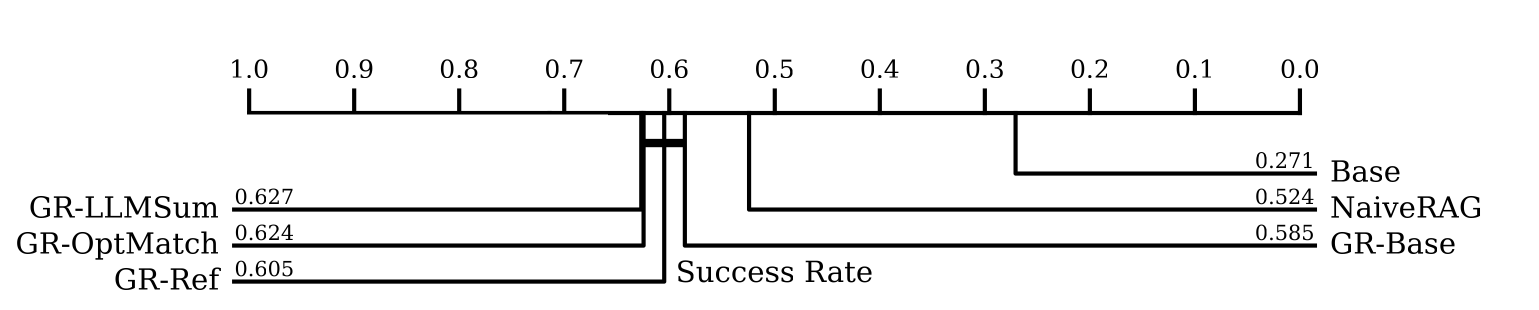}
    \caption{Critical Difference Diagram for Overall Success Rates (both Technical Validation and Intent Validation). Methods connected by horizontal lines show no statistically significant difference based on pairwise McNemar tests.}
    \label{fig:overall_success}
\end{figure}

The CD diagrams in Figures~\ref{fig:tv_success} and~\ref{fig:overall_success} reveal distinct performance clusters that emerge when statistical significance is assessed across all methods simultaneously. For both technical validation and overall success metrics, the pairwise McNemar test results indicate that GR-Base, GR-OptMatch, GR-LLMSum, and GR-Ref form a statistically equivalent cluster, despite some methods within this group showing significant differences in direct pairwise comparisons. This clustering phenomenon occurs because the multiple-comparison adjustment reveals that, while individual methods may differ significantly when compared in isolation, their relative performance differences become less pronounced when evaluated within the complete experimental landscape.

\section{Discussion}
\label{sec:discuss}
This section discusses the main findings and their implications. 

\paragraph{Insufficient Parametric Knowledge}
\label{ssec:insuff_param}
The most significant finding is the dramatic performance improvement achieved through structured knowledge representation. While baseline LLM generation achieved only 27.1\% overall success, our GR-LLMSum reached 62.6 \%. This leap is not an incremental improvement; rather, it demonstrates that IaC generation is precisely the kind of knowledge-intensive task where the parametric knowledge of LLMs proves fundamentally insufficient \cite{huang2025survey, gao2023retrieval}. It suggests that an LLM’s internal representation of a formal domain like IaC is a form of lossy compression. The model learns statistical likelihoods from code it has seen, but it struggles to store and reliably reproduce the rigid, symbolic rules of a provider's schema, a known challenge for LLMs in domain-specific languages \cite{kon2024iac, pan2024unifying}.
A knowledge graph compels the model to anchor its probabilistic outputs in a deterministic source of truth \cite{hogan2021knowledge, pan2024unifying}. This aligns with recent findings~\cite{wang2025large} showing that integrating knowledge graphs can reduce hallucination and supply accurate, domain-specific information to LLMs. In formal domains like IaC, this implies that reliability depends not on scaling models further, but on building architectures that incorporate external, grounded reasoning \cite{kon2024iac, pan2024unifying}.

\paragraph{Correctness-Congruence Gap}
Perhaps the most revealing finding is the divergence between technical validation and intent validation. While our best method, GR-OptMatch, increased the technical success rate to 84.2\%, the intent validation rate plateaued at 72-76\% across all RAG-based methods. Our previous analysis (Section \ref{ssec:insuff_param}) showed that these persistent intent failures are not simple mistakes; they are predominantly complex congruence issues, such as incomplete multi-resource configurations and misaligned operational parameters. This difference between achieving technical script validity and fulfilling true user requirements reveals what we term the "Correctness-Congruence Gap". 
\begin{itemize}
     \item \textbf{Correctness} is adherence to syntactic and schematic rules \cite{zan2022large, kon2024iac, tambon2025bugs}. It is a closed-world problem, verifiable against a formal specification, such as a schema, in our case, modeled as a knowledge graph. Our findings show this is largely solvable through structured knowledge injection.
     
     \item \textbf{Congruence} is the alignment of generated code with the user's unstated goals, operational context, and implicit best practices \cite{ouyang2022training, chen2021evaluating}. This is an open-world problem requiring architectural reasoning: inferring dependencies~\cite{yu2024codereval}, anticipating security implications~\cite{tambon2025bugs}, and understanding the system's functional purpose~\cite{zhong2024can}.
\end{itemize}

The plateau exists because our methods are optimized to solve for Correctness, but the remaining failures are almost entirely problems of Congruence. This suggests that current models can function as excellent "IaC coders" when given a detailed specification and sufficient context~\cite{tan2024prompt} but remain poor "architects" when given a high-level objective \cite{yu2024codereval}. Bridging this gap may require new approaches that explicitly model design patterns and constraints, moving beyond the current capabilities of our Graph RAG.

\paragraph{Context Optimization}
The unexpected failure of comprehensive dependency modeling (GR-Ref) on simple tasks revealed a "cognitive overload" effect: including all referenced resources, though beneficial for complex configurations, degraded performance on simpler ones by introducing distracting or irrelevant information. This suggests that simply adding more context is not always beneficial, a finding consistent with prior studies in traditional RAG, which show that noisy or excessive context can harm performance \cite{gao2023retrieval, huang2025survey}. Our findings extend this observation to IaC generation with Graph RAG architectures, prompting what we term the "Principle of Optimal Context": effective systems should not aim to maximize context, but to optimize it.

\subsection{Implications for Researchers}
The findings detailed above reveal several fundamental challenges that can guide future research directions for automated IaC generation.

\paragraph{Addressing the Correctness-Congruence Gap} 
The "Correctness-Congruence Gap" is the most critical theoretical challenge revealed by this research. The fact that technical correctness can be systematically solved while intent alignment remains a persistent plateau implies two primary directions for researchers:

\textbf{Move Beyond Translation to Architectural Reasoning:} The gap shows that treating IaC generation as a simple natural-language-to-code translation task is insufficient. The real challenge lies in bridging the abstraction levels between high-level user goals (the "problem space") and low-level, interconnected resource specifications (the "solution space"). Future research should focus on developing models capable of performing architectural reasoning.

\textbf{Develop Nuanced Evaluation for Intent:} The binary pass/fail of the intent validation framework masks the complexity of the problem. A generated configuration that is 90\% aligned with user intent is currently treated the same as one that is 10\% aligned. Researchers must develop more nuanced evaluation frameworks that can measure degrees of congruence, recognize partially correct solutions, and account for the validity of multiple different implementation strategies for the same high-level goal.

\paragraph{Advancing Knowledge Engineering}
The research demonstrates that knowledge representation significantly impacts generation quality, but reveals trade-offs between comprehensiveness and precision. The finding that GR-Ref improved complex configurations while degrading simple ones suggests that optimal knowledge injection must be context-aware and adaptive. The goal should not be to maximize the amount of retrieved information, but to tailor its complexity and granularity to the specific query.

Furthermore, the plateau in semantic enhancement effectiveness (GR-LLMSum showed no improvement over GR-OptMatch) indicates that retrieval bottlenecks may not lie in description quality but in the fundamental challenge of mapping user intent to relevant knowledge. This suggests that future research should investigate query understanding and intent decomposition rather than focusing solely on improvements to knowledge representation.

\begin{tcolorbox}[colback=black!5, colframe=gray!80, title=Academic Insight Summary]
We demonstrate how the knowledge representation improves the quality of LLM-based IaC generation.
\end{tcolorbox}

\subsection{Implications for Practitioners}

\paragraph{The Reality of Current LLM-based IaC Generation Limitations.} 
While the research demonstrates substantial improvements through structured knowledge injection, practitioners must understand that even the best-performing approach (GR-LLMSum) achieves only 62.6\% overall success. Organizations considering automated IaC generation must plan for these failure rates.

Even when the generated code is technically valid and deployable, it may not fulfill the user's actual requirements (intent failures). The most critical risk is not a script that fails, but a script that deploys the wrong infrastructure. Organizations must implement validation processes that go beyond technical correctness to ensure that the generated infrastructure aligns with business requirements, security policies, and operational constraints.

\paragraph{The Trade-off between Generality and Domain Specificity.} The significant improvement from 27.1\% to 62.6\% overall success through domain-specific knowledge injection demonstrates clear benefits of specialization, but also reveals the substantial effort required to achieve reliable automated generation. Organizations adopting LLM-based IaC generation must accept that reliable IaC generation requires a dedicated, structured knowledge source. This is not a one-time setup but a long-term operational commitment to maintaining an authoritative knowledge graph of your cloud provider's schemas and your organization's specific infrastructure patterns.

\paragraph{Operational Implications of IaC development} 
The IaC workflow should be designed around the assumption that the first-pass generation is a draft. The process should be: generate the code, automatically validate it for technical correctness, and then pass it to an engineer for congruence review and final approval. The \~40\% overall failure rate should be treated as a baseline for workflow planning, not as an exception.

\paragraph{Future Perspectives on Infrastructure Engineering Roles.} 
LLM-driven IaC generation will not eliminate the need for infrastructure engineers, but it likely will transform their role. Rather than focusing on low-level implementation details, engineers will shift toward high-level specification, validation, and system thinking. In this new paradigm, a key skill is the ability to bridge the Correctness-Congruence Gap, ensuring that technically correct code also aligns with the user's true intent, operational context, and architectural constraints. This involves specifying high-level intent in a form that LLMs can understand, validating whether the generated code satisfies functional, security, and business requirements, and curating the domain-specific knowledge that powers these generative systems. 
Rather than eliminating manual work, these systems augment engineers' capabilities, freeing them from writing boilerplate code so they can focus on architectural reasoning. Practitioners must become experts in working with AI-generated code, understanding its failure modes, and designing robust validation processes that ensure its safe and effective use.

\begin{tcolorbox}[colback=black!5, colframe=gray!80, title=Practitioner Insight Summary]
IaC generation systems are not ready to be fully automated inside the IaC development industrial workflow. The infrastructure engineers still need to validate the infrastructure they want to deploy, and IaC generation systems can support their work.
\end{tcolorbox}

\section{Limitations and Threats to Validity}
\label{sec:threats}
Several limitations affect the validity and generalization of this research, spanning both the evaluation methodology and the experimental design.

\textbf{Construct Validity Threats.} The primary threats in this study relate to how IaC correctness is defined and measured: \textbf{(i)} The error taxonomy, systematically developed through qualitative coding, represents one possible categorization of generation failures.  Alternative taxonomies might reveal different patterns or emphasize different aspects of model behavior, affecting the interpretation of why specific knowledge
injection approaches succeed or fail. In addition, the evaluation approach assesses the functional correctness of the IaC without considering other aspects as maintainability, readability, or efficiency. \textbf{(ii)} The intent validation framework, despite extensive refinement, ultimately relies on human-authored specifications of what constitutes correct infrastructure.

\textbf{Internal Validity Threats.} Methodological decisions and data processing choices may have systematically influenced the results.
\textbf{(i)} The manual correction of OPA could introduce potential bias; to limit it, the manual correction was done by experts in Terraform. 
\textbf{(ii)} The prompt dataset introduces potential sampling limitations. The linguistic and technical distribution of the benchmark dataset inherently shapes which aspects of infrastructure generation capability are measured, potentially overlooking important use cases or domain-specificities.

\textbf{External Validity Threats.} The primary limitations relate to scope constraints and temporal validity: \textbf{(i)} The experimental scope is constrained to Terraform configurations for AWS infrastructure, representing only one IaC tool and cloud provider. While syntactic differences exist between tools and providers, the foundational infrastructure management concepts, resource dependencies, state management, and configuration provisioning, remain consistent across the IaC landscape \cite{nasiri2024towards, hassan2024state}
\textbf{(ii)} The research focuses exclusively on a single large language model (GPT-4o) to isolate the effects of knowledge injection methods from model-specific variations. This design choice could limit broader applicability with an alternative LLM. 
\textbf{(iii)} The benchmark dataset comprises 458 human-curated scenarios, which may
not be representative of real-world infrastructure authoring tasks. 
(\textbf{iv)} The evaluation timeframe represents a snapshot of LLM capabilities and cloud provider schemas. As they evolve, the relative effectiveness of different knowledge-injection approaches may change, potentially affecting the long-term validity of the findings.

\section{Conclusions and Future Works}\label{chapter:conclusions}
In this research, we studied the effects of configuration knowledge in the LLM-based IaC generation.  We first enhanced an existing benchmark framework used for assessing LLMs for Terraform IaC generation. Next, we developed a two-dimensional taxonomy comprising 15 error types to categorize errors in LLM-generated Terraform code. We observed that the LLM's inaccurate or incomplete knowledge about Terraform configuration was a significant cause of errors in generated IaC scripts. Hence, we implemented various knowledge injection techniques and conducted a comprehensive experimental evaluation. Our findings show that structured knowledge representation, particularly through graph-based systems such as Graph RAG, significantly improves technical correctness. Graph RAG achieved a technical validation success rate (i.e., generation of valid IaC code) of 80.3\%, surpassing both Naive RAG (70.2\%) and the baseline with no configuration injection (37.2\%). The enhancements to Graph-RAG further improved technical validation, increasing success rates to 84.2\%. However, the study also reveals a nuanced trade-off: structured knowledge excels at technical validation but does not significantly improve intent alignment (i.e., alignment between the user's high-level goals for the infrastructure and the generated infrastructure).

In future work, we plan to extend our overall framework, including the taxonomy and the IaC generation framework, to support other IaC languages. We also aim to leverage additional sources of IaC knowledge, including GitHub repositories and Stack Overflow, to improve code generation accuracy and develop code repair strategies.

\bibliographystyle{ACM-Reference-Format}
\bibliography{sample-base}

\end{document}